\documentclass[10pt]{article} 
\usepackage{rlc}

\usepackage{amssymb}            
\usepackage{mathtools}          
\usepackage{mathrsfs}           
\mathtoolsset{showonlyrefs}     
\usepackage{graphicx}           
\usepackage{subcaption}         
\usepackage[space]{grffile}     
\usepackage{url}                

\usepackage{amsmath}
\usepackage{multirow}
\usepackage{multicol}
\usepackage{lscape}
\usepackage{array}

\title{Is Exploration All You Need? Effective Exploration Characteristics for Transfer in Reinforcement Learning}

\author{Philip S. Thomas  \\
    pthomas@cs.umass.edu \\
    College of Information and Computer Sciences\\
    University of Massachusetts
    \And
    Glen Berseth \\
    glen.berseth@mila.quebec\\
    Mila, Universit\'e de Montr\'eal \\
    CIFAR Canada AI Chair}

\begin{document}

\maketitle

\begin{abstract}

In deep reinforcement learning (RL) research, there has been a concerted effort to design more efficient and productive exploration methods while solving sparse-reward problems.
These exploration methods often share common principles (e.g., improving diversity) and implementation details (e.g., intrinsic reward). 
Prior work found that non-stationary Markov decision processes (MDPs) require exploration to efficiently adapt to changes in the environment with online transfer learning.
However, the relationship between specific exploration characteristics and effective transfer learning in deep RL has not been characterized. 
In this work, we seek to understand the relationships between
salient exploration characteristics and improved performance and efficiency in transfer learning. 
We test eleven popular exploration algorithms on a variety of transfer types---or ``novelties''---
to identify the characteristics that positively affect online transfer learning.
Our analysis shows that some characteristics correlate with improved performance and efficiency across a wide range of transfer tasks, while others only improve transfer performance with respect to specific environment changes.
From our analysis, make recommendations about which exploration algorithm characteristics are best suited to specific transfer situations.

\end{abstract}

\section{Introduction}
\label{sec:introduction}

Modeling the world as a stationarity Markov decision process is fundamental to the understanding and advancement of reinforcement learning (RL). 
However, many real-world problems used to motivate RL research, such as robotics~\citep{rlroboticssurvey}, autonomous driving~\citep{kiran2021surveydriving}, power distribution~\citep{powerrlsurvey}, and language preferences~\citep{casper2023rlhf}, are in fact non-stationary. 
Although robustness can accommodate small amounts of nonstationarity, many environmental changes---or ``novelties''---require agents to adapt. 
Novelties are sudden changes to the observation space or environment state transition dynamics that occur at inference time that are unanticipated (or unanticipatable) by the agent during training~\citep{boult2021towards,balloch2022novgrid}.
When there is a novel change, the RL agent's converged policy is no longer reliable and the agent can become ineffective or even make catastrophic and harmful mistakes.

One method for adaptation in RL is transferring prior knowledge to the new environment, known as task transfer or transfer learning~\citep{taylor2009transfer}. 
In real-world scenarios, transfer may need to be done in-situ or \textit{online}, referred to as online task transfer (OTT)~\citep{zhan2015online}.

Mirroring findings in biology animal behavior~\citep{reale2001temperament}, prior work has found that exploration can improve the RL agent performance and efficiency in transfer~\citep{taylor2009transfer,silver2013lifelong,langley2022agents}. 
In deep reinforcement learning research, there has been widespread interest to conceive more efficient and rewarding exploration methods.  
In theory, exploration designed for stationary RL can enable agents to experience and adapt to environment novelties with no fundamental changes~\citep{schmidhuber1991curious,Schmidhuber91apossibility,chentanez2004intrinsically}.
In spite of this, exploration algorithms designed to improve the exploration-exploitation trade-off of RL in stationary environments have not been comprehensively analyzed for their effect on efficient online task transfer. 

In this research, we ask the question: 
\textbf{which attributes of traditional exploration algorithms are important for efficient transfer in RL?}

In this paper we investigate the effectiveness of exploration methods designed for stationary RL problems on non-stationary OTT problems. 
We identify important characteristics of reinforcement learning exploration algorithms, then systematically examine the within- and between-class relationships of these characteristics with a variety of OTT problems and environments. 
We then report on the results of a first-of-its-kind large-scale experiment across eleven exploration algorithms on five distinct two-environment online task transfer problems. 

Our findings indicate that agent performance on transfer problems is indeed sensitive to the choice of exploration algorithm, specifically that:
(1) exploration principles of \textit{explicit diversity} and \textit{stochasticity} are the most consistently positive exploration characteristics across a variety of novelty and environment types, 
(2) source task convergence efficiency in discrete control tasks is inversely correlated with target task adaptive efficiency, while the opposite is true in continuous control, 
(3) transfer problems in continuous control environments are more negatively impacted by time-dependent exploration methods compared to discrete control, and 
(4) the relative importance of exploration characteristics like explicit diversity varies with novelty type.

\section{Preliminaries}

Reinforcement learning typically models an environment as a stationary Markov decision process (MDP): $M=\langle \mathcal{S}, \mathcal{A}, R, P, \gamma\rangle$, 
where $\mathcal{S}$ is the space of environment states, $\mathcal{A}$ is the space of actions, $R: \mathcal{S} \times \mathcal{A} \rightarrow \mathbb{R}$ is the function that maps states and actions to a scalar reward, $P: \mathcal{S} \times \mathcal{A} \rightarrow \mathcal{S}$ is the transition function between states, and $\gamma$ is the discount factor of future reward. 
The learning task in RL is to learn a policy $\pi(a \vert s)$ that, for a given state, selects the action that maximizes expectation of discounted future reward.
The policy is related to the reward through the notions of state value $V_{\pi}(s)$ and state-action value $Q_{\pi}(s,a)$ in the Bellman Expectation Equations~\citep{sutton2018reinforcement}:
%
$
    V_{\pi}(s) = \sum_{a \in \mathcal{A}} Q_{\pi}(s, a) \pi(a \vert s)$
where
$
    Q_{\pi}(s, a) = R(s, a) + \gamma \sum_{s' \in \mathcal{S}} P_{s,s'}^a V_{\pi} (s')
$.

The policy gradient theorem~\citep{sutton2018reinforcement} forms the basis of how policy gradient algorithms, such as actor-critic, can find the optimal policy parameters by maximizing the value of on-policy decisions: 
%
$
\nabla_\theta J(\theta) = \nabla_\theta \mathbb{E}_{\pi} \left[
 Q_{\pi}(s, a) \pi_\theta(a \vert s) \right] 
\propto \mathbb{E}_{\pi} \left[ Q_{\pi}(s, a) \nabla_\theta \pi_\theta(a \vert s) \right]
$
%

\paragraph{Exploration}
In reinforcement learning, agents cannot greedily pursue future reward; to find an optimal policy they must trade off exploration with exploitation~\citep{sutton2018reinforcement}. 
Exploitation encourages the agent to repeat behaviors that yielded rewards in the past, while exploration helps the agent discover opportunities to capture even more rewards than in previous experience. 
Exploration is also to cope with non-smooth learning challenges such as local minima and credit assignment challenges that result from sparse or time-varying rewards. 
However, there is no one convention for how to execute this trade-off; RL algorithms only define how the policy should assign value and learn, assuming that greedy exploitation is enough. This means that there are many different ways that RL exploration algorithms are designed, motivated, and integrated into the learning process. 

\paragraph{Transfer Learning}
When a parameterized model has knowledge from a prior problem, it is natural to attempt to update it using data from a new target problem. 
However, sampling data from distribution could induce catastrophic inference~\citep{mccloskey1989catastrophic}, causing the agent to transfer little, if any, of its previous model when learning the new task. 
\textit{Transfer learning}~\citep{zhu2023transfer} seeks to overcome this limitation when transferring the source model to a new target environment that differs in nontrivial ways---what are referred to as ``novelties''~\citep{boult2021towards,balloch2022novgrid}.

In transfer learning for supervised and semi-supervised learning, the problem is usually modeled as a sequence of discrete domains and tasks. 
\textit{Domain} is defined as~$\mathcal{D} = \{\mathcal{X}, p(X)\}$, where $p(X)$ is the marginal distribution over the input dataset $X$ sampled from the input space $\mathcal{X}$, and \textit{learning task} is defined as~$\mathcal{T} = \{\mathcal{Y}, p(Y \vert X)\}$, where $p(Y \vert X)$ is the distribution of the outputs $Y$ from the output space $\mathcal{Y}$ given the inputs~\citep{pan2009survey}. 
In transfer learning, there is a \textit{source} task $\mathcal{T}_s$ and domain $\mathcal{D}_s$ for which a model is originally optimized, and a \textit{target} task $\mathcal{T}_t$ and domain $\mathcal{D}_t$ on which the performance of that model will be measured. 

Following this discrete approach, we model the transfer process for deep reinforcement learning as a transition from \textit{$MDP_\mathrm{source}$} and \textit{$MDP_\mathrm{target}$}. 
As the environment is a decision process to be sampled through interaction instead of a presampled set, the formulation of $\mathcal{D}$ and $\mathcal{T}$ is distributed according to the sampling procedure, which is non-i.i.d as it is a function of the policy and, by association,  exploration method.
Therefore, the learning domain becomes $\mathcal{D} = \{\mathcal{S}, P\left(s^{\prime}, r \mid s, a\right) \}$.
The learning task, on the other hand, is a function of the approach; for example, in model-free RL like Q-learning the task becomes~$\mathcal{T} = \{\mathcal{A}, Q(S_t, A_t)\}$. 


This work focuses specifically on the \textit{ ``online task transfer''}~\citep{zhan2015online} problem domain in RL. In online task transfer, RL agents are trained on a source task then are asked to adapt to a target task associated with a new MDP while being evaluated in that new target MDP.
This process occurs ``online,'' where the agent suddenly experiences the novel change in the MDP without any prior knowledge of the target MDP or when the transfer will occur. 

\begin{figure}[t]
    \begin{center}
        \includegraphics[width=\textwidth]{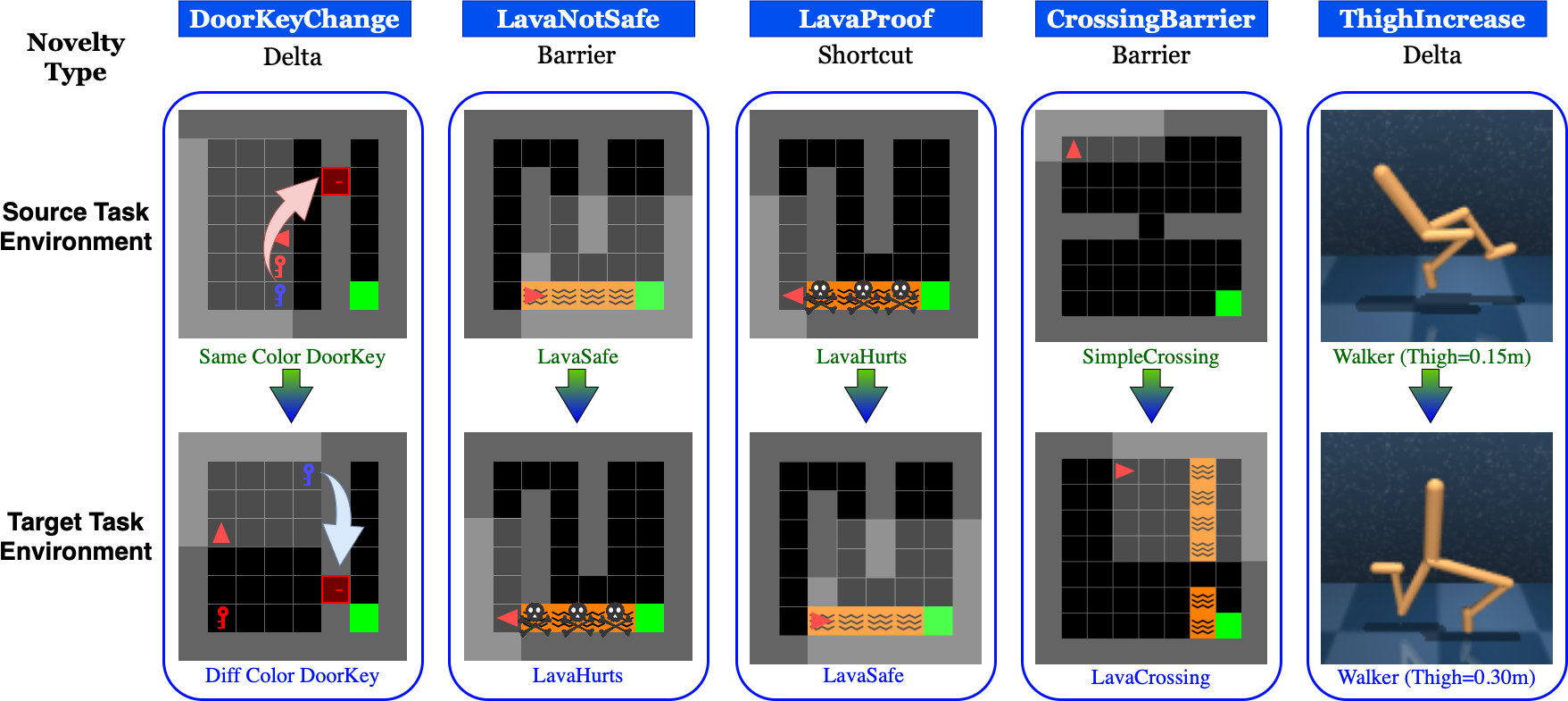}
    \end{center}
    \caption{Environments and novelties used to evaluate the exploration algorithms and their characteristics. 
    This shows the mixture novelties and discrete and continuous control environments.} 
    \label{fig:envs}
\end{figure}

\section{Characterizing Exploration Methods}

As already noted, there are many ways to incorporate exploratory signals into reinforcement learning. 
Thus, it can be useful to characterize the way in which any given RL process is influenced by exploration and how it might choose between alternate behaviors. 
Looking at these differing influences and behaviors, and drawing inspiration from existing taxonomies of exploration methods~\citep{ladosz2022exploration},  we break down the characteristics of exploration algorithms into broadly three categories: \textit{exploration principle}, \textit{temporal locality}, and \textit{algorithmic instantiation}. 

\textbf{Exploration principle}
    characterizes an agent's behavior beyond greedy maximize reward. This can be thought of as the ``philosophy'' underlying the exploration method.
We consider three exploration principles. 
(1) The integration of added \textit{stochasticity} into the learning process.
There are many ways to use stochasticity in exploration, whether by injecting random noise into the input or some intermediate weight layer, using a stochastic policy, distilling a random network, or simply selecting random actions. 
(2) \textit{Explicit diversity} over the different random variables in the process.
Explicit diversity methods ensure that the a greedy process doesn't lead the agent into stale transitions, effectively encouraging models to experience all parts of the domain and task equally.
(3) Having a \textit{separate objective} in addition to greedy pursuit of reward. 
Lastly, methods with a separate objective identify an objective that complements the flaws of greedy reward maximization and either alternates between or combines these two objectives. 

\textbf{Temporal locality}
characterizes the relationship an exploration algorithm has with time. 
Most exploration algorithms are designed in an effort to be adaptive to the different learning needs of an agent at different points in learning, making them time-dependent even if only implicitly. 
Algorithms with short-term or temporally \textit{local} characteristics implement adaptive behavior as a function of how agent and environment properties evolve time step to time step or episode to episode. 
Algorithms with long-term temporally \textit{global} characteristics instead look at agent and environment properties recorded or aggregated across the entire learning problem and direct exploration based on trends in global properties or comparing global properties with the current agent, environment, or learning state.
As in the other two category there are algorithms like Never Give Up~\citep{Badia2020Never} where a core contribution is the combination of both temporally local and global thinking. 
However, there are some truly \textit{time-independent} exploration methods. 
Similarly to other characterizations~\citep{sutton2018reinforcement} of exploration methods as ``directed'' or ``undirected,'' time-independent methods avoid the temporal locality question altogether, counteracting greedy behavior by passively altering the learning process or in the agent architecture. 
This becomes critical to evaluation of exploration in transfer applications because online task transfer induces a temporal shift, both globally and locally. 
All time-dependent methods can struggle or improve relative to a baseline by adapting exploration through the transfer process.

\textbf{Algorithmic instantiation} 
characterizes the mechanism within the reinforcement learning process that alters the typical greedy mechanisms.
Fundamentally the reinforcement learning process can be thought of as cycle with two directions: ``forward,'' where the agent interacts with the environment, receives reward, and collects samples for learning, and ``backward,'' where the agent's models are updated according to the update function based on reward and a loss is calculated and applied based on the reward and update. 
We consider three means of algorithmic instantiation. 
(1) Exploration-based \textit{environment sampling}.
Different means of sampling non-greedily, for example randomly or for explicity diversity, affect the forward process, making the data distribution more amenable to finding the optimum. 
(2) A modification of the \textit{update function}.
Modifying the reinforcement update process, including but not limited to the loss function, affects the forward process by propagating incentives to the agent that are not greedy reward maximization. 
(3) The addition of an \textit{intrinsic reward}. 
Intrinsic motivation is a quality of exploration methods that incentivize visitation of sub-optimal transitions by reweighting the rewards experienced by the agent at those transitions. Intrinsic reward is unique because it is not definitively part of the forward or backward processes: exploration can just as easily sample states and actions according to an intrinsic reward and alter the agent update with intrinsic reward. 

We organize the exploration methods studied in this work in relation to these categories, as summarized in Table~\ref{tab:characteristics} and Appendix~\ref{app:chars}.

\begin{table}[tb!]
\centering
\footnotesize
\begin{tabular}{>{\raggedright\arraybackslash}p{0.2\textwidth}>{\raggedright\arraybackslash}p{0.3\textwidth}>{\raggedright\arraybackslash}p{0.4\textwidth}}
\hline
\textbf{Categories} & \textbf{Characteristics} & \textbf{Example Algorithms} \\
\hline \hline
\textbf{Algorithmic Instantiation} & 
\begin{itemize}
    \item Environment Sampling
    \item Update Function
    \item Intrinsic Reward
\end{itemize} &  
\begin{itemize}
    \item DIAYN
    \item NoisyNets
    \item RND, REVD, RISE, RE3, RIDE, ICM, NGU, DIAYN, GIRL
\end{itemize} \\
\hline
\textbf{Exploration Principle} & 
\begin{itemize}
    \item Stochasticity
    \item Explicit Diversity
    \item Separate Objective
\end{itemize} & 
\begin{itemize}
    \item NoisyNets, DIAYN
    \item RND, REVD, RISE, RE3, RIDE, NGU, DIAYN
    \item RND, RIDE, ICM, NGU, GIRL
\end{itemize} \\
\hline
\textbf{Temporal Locality} & 
\begin{itemize}
    \item Global 
    \item Local
    \item Time Independent
\end{itemize} & 
\begin{itemize}
    \item RND, ICM, RE3, NGU, GIRL
    \item REVD, RIS, RIDE, NGU
    \item NoisyNets, DIAYN
\end{itemize} \\
\hline
\end{tabular}
\caption{This table lays out our decomposition of exploration algorithms into three major categories---exploration principle, algorithmic integration, and temporal locality---with three core characteristics in each.
The algorithms we list here are the algorithms we evaluate as described in Section~\ref{sec:algos}
Algorithms are described in detail in the Appendix.}
\label{tab:characteristics}
\end{table}

In this work, we focus primarily on the effects of exploration principle and temporal locality characteristics. 
This is primarily because our results showed that different exploration principles and temporal localities had a critical and varied impact on different transfer scenarios, while our results did not show significant trends associated with varying algorithmic instantiations. 
We further examine possible reasons for this in Appendix~\ref{app:chars}.

\begin{figure}[t]
    \begin{center}
        \includegraphics[width=0.75\textwidth]{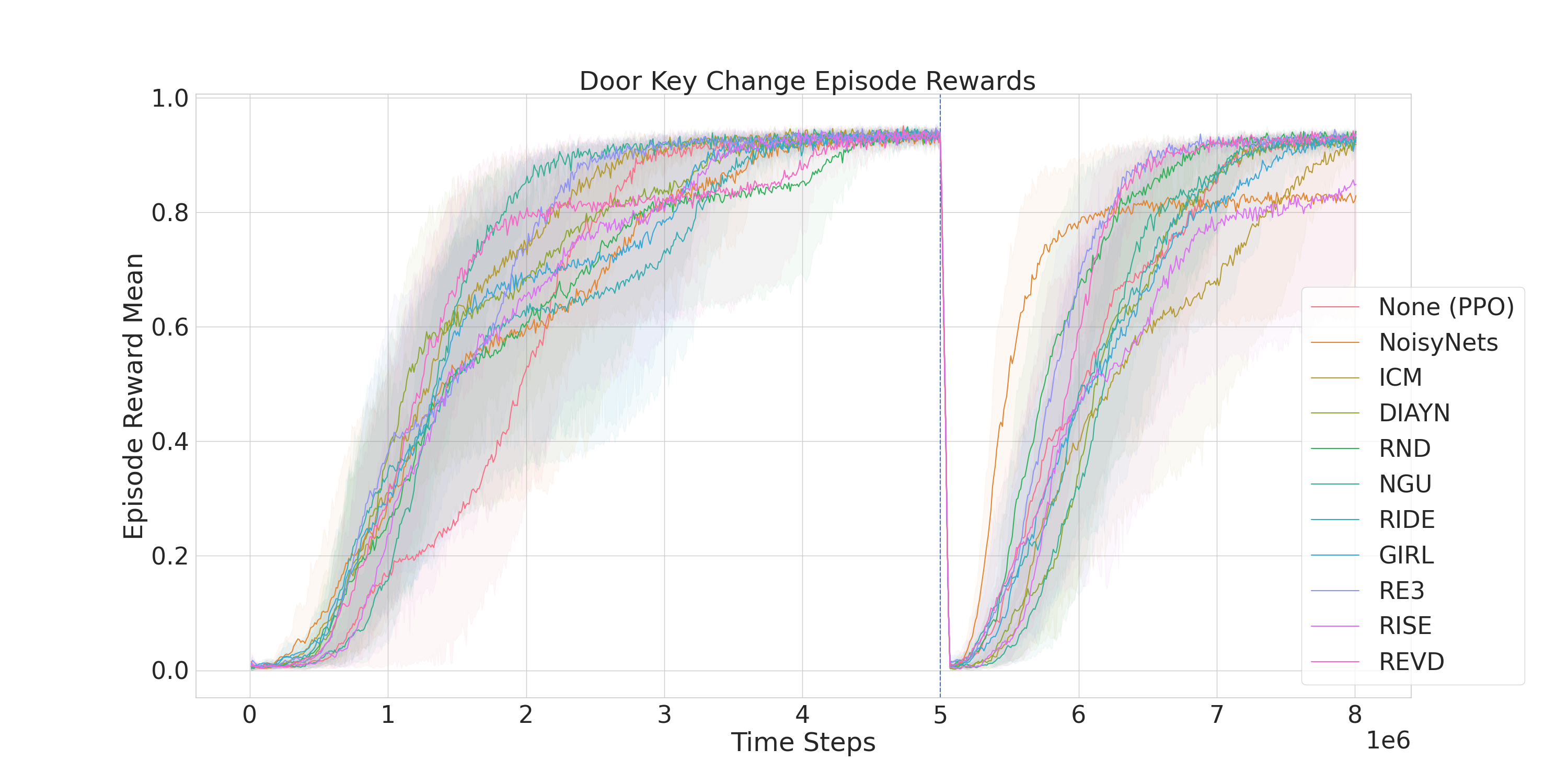}
    \end{center}
    \caption{Full learning and adaptation process of all eleven RL exploration algorithms on the DoorKeyChange novelty problem from NovGrid~\citep{balloch2022novgrid}. 
    11 different RL agents first learn a task assuming a stationary MDP. 
    The rate of learning at this stage is {\em convergence efficiency}.
    At time step 5,000,000 novelty is injected into the environment, transfering from $MDP_\mathrm{source}$ to $MDP_\mathrm{target}$, and often causing a performance drop-off. 
    The algorithms then recover their past performance as they learn the new world transition dynamics. 
    The rate of learning at this stage is {\em adaptive efficiency}.
    The maximum episode reward is the {\em final adaptive performance}, which may not always be as high as pre-novelty performance.
    } 
    \label{fig:full_group_doorkey}
\end{figure}

\section{Experimental Setup}

With our experiments we seek to address the following questions:
    (1)
    Are there characteristics of existing exploration algorithms that empirically demonstrate greater suitability for online task transfer overall?
    (2)
     How do specific exploration characteristics impact transfer performance in discrete and continuous environments  differently? 
    (3)
    Do specific exploration characteristics lend themselves well to specific types of environment change?

\subsection{Exploration Algorithms}\label{sec:algos}

For our assessment, we focus on model-free, on-policy deep policy gradient methods that apply to a variety of reinforcement learning tasks. 
Specifically, we use proximal policy optimization (PPO)~\citep{schulman2017proximal}, a high-performing actor-critic policy gradient method, as the algorithmic backbone of all the exploration methods we test. 
On-policy actor-critic methods such as PPO are more versatile than off-policy methods, which only apply to a subset of RL problem formulations. 
For example, methods like Deep Q-Networks~\citep{mnih2015human} only apply to problems with discrete action spaces and methods like Soft-Actor Critic~\citep{haarnoja2018soft} and Deep Deterministic Policy Gradients~\citep{lillicrap2019continuous} only work in continuous control environments. 
Additionally, off-policy methods are very sensitive to the management of an experience replay buffer for successful learning~\citep{mnih2015human}, which becomes significantly more complex when adapting online
because hyperparameters such as how often the experience replay buffer should be reset become potential confounding variables.
In an effort to control as many independent variables as possible and focus our investigation on exploration, we only consider the PPO algorithm for our investigation.

We select 11 popular exploration algorithms compatible with PPO and our RL settings that represent a broad sampling of exploration principles and algorithmic implementations.
Those algorithms are Random Network Distillation (RND)~\citep{burda2018exploration}, Intrinsic Curiosity Module (ICM)~\citep{pathak2017icm}, Never Give Up (NGU)~\citep{Badia2020Never}, Rewarding Impact-Driven Exploration (RIDE)~\citep{raileanu2019ride}, Renyi State Entropy Maximization (RISE)~\citep{yuan2022renyi}, Rewarding Episodic Visitation Discrepancy (REVD)~\citep{yuan2022rewarding}, enerative Intrinsic Reward Learning (GIRL)~\citep{yu20girl}, Parameter Space Noise for Exploration (NoisyNets)~\citep{plappert2018parameter}, and ``online'' Diversity Is All You Need (DIAYN)~\citep{eysenbach2018diversity}. 

These algorithms were selected because 
they reflect a good coverage of the characteristics we wish to compare in different OTT problems.
Table~\ref{tab:characteristics} shows how the algorithms relate to exploration characteristics and descriptions of the algorithms can be found in Appendix~\ref{app:algos}.
Our implementation of these algorithms is based on the Stable-Baselines3~\citep{raffin2021stable} and RLeXplore libraries,\footnote{https://github.com/RLE-Foundation/RLeXplore} 
which we modify and expand for the purposes of our investigation.

\subsection{Learning Environments and Transfer Tasks}

To experiment with online transfer, agents are trained to convergence in one environment (the source task), and then a novelty is applied to create the target task. 
The agent must recover its performance during online execution in the target environment.
We run our experiments in two environments, the NovGrid~\citep{balloch2022novgrid} and Real World Reinforcement Learning suite~\citep{Dulac-Arnold2021} transfer learning libraries. 
NovGrid is a specialization of the MiniGrid~\citep{MinigridMiniworld23} environment designed to promote experimentation in online transfer.
Specifically, NovGrid sets up learning scenarios then injects a novelty---changing the transition dynamics---at a time that is unknown to the agent.
For example, the \texttt{DoorKeyChange} novelty is to change which colored key opens which door;
The \texttt{LavaSafe} novelty causes normally lethal lava to become non-damaging; and \texttt{LavaHurts} causes lava that was initially safe to become lethal. 
For our experiments we use these three novelty environments.
\texttt{DoorKeyChange} represents a novelty in which one solution trajectory is replaced by a new, different solution trajectory that is approximately the same length and complexity.
\texttt{LavaSafe} represents a novelty in which the solution trajectory in the new environment is substantially shorter, cheaper, and less complex.
\texttt{LavaHurt} represents a novelty in which the solution trajectory in the new environment is substantially longer, more expensive, and more complex. 
These capture the three main types of novelties~\citep{balloch2022novgrid}: deltas, shortcuts, and barriers.

We allowed the algorithms to run until the majority of runs on all algorithms converged---varying from 1-20 million steps---before the novelty was injected. 
We tuned the hyperparameters of the algorithms on the novelty-free \texttt{DoorKey} environment for use with the NovGrid environments, maximizing convergence in the source environment so as to both (a)~find hyperparameters that were well suited to the specific source MDP and (b)~to help ensure convergence on the source task. The details of the hyperparameter tuning is in Appendix~\ref{app:hyperparams}.  

We also evaluated the adaptation performance of algorithms in continuous control environments using the Real World Reinforcement Learning suite~\citep{Dulac-Arnold2021}. 
We tuned the hyperparameters of our algorithms on the Cartpole-Swingup environment by changing the pole length, which maintains the same approximate difficulty of the target task.
We evaluate OTT performance on the more complex Walker2D environment by varying thigh length from 0.15 meters to 0.3 meters.
See Figure~\ref{fig:envs} for illustrations of the environments and novelties.

\subsection{Metrics for Online Task Transfer}

Our metrics are focused on measuring learning efficiency and performance, motivated by the desire to minimize the number of environment interactions required to learn good policies.
These metrics include: 

\textbf{Convergence efficiency}: The number of environment steps necessary for the agent to reach 95\% of maximum performance on the original source task. 

\textbf{Adaptive efficiency}: The number of environment steps necessary for the agent to reach 95\% of maximum performance on the target task.

\textbf{Final [adaptive] performance}: The total episodic return of an agent that converged on a task. Final performance can be measured on either task, but when called ``final \textit{adaptive} performance'' it specifically refers to the performance on the target task at some point $K$ after which the majority of the runs of all algorithms converged. In our experiments, we set $K$ for each task to be the mean adaptive efficiency of the algorithm with the worst adaptive efficiency.  

\textbf{Transfer Area Under the Curve} (Tr-AUC):
Inspired by the performance ratio of \cite{taylor2009transfer}, we designed Tr-AUC as a novelty-agnostic measure of the overall transfer performance as a function of both the source and target task:
\begin{equation}
\text{Tr-AUC} = \frac{1}{2} \left(\max(r_{\text{S}})+\frac{1}{K} \sum_{i \in K} r_{i,\text{T}} \right)
\label{eq:trauc}
\end{equation}
%
where $\max(r_{\text{S}})$ refers to the final performance on the source task and and the summation over $r_\text{T}$ gives accumulated adaptive performance until the final adaptive performance point on the target task.  
The benefit of and principle behind Tr-AUC is that it balances rapid and effective adaptation with prior task performance. 
Specifically, we noticed early in our experiments that many of the best performing agents on the target task metrics were the algorithms that performed poorly or failed to converge on the source task. 
This is because one of the core challenges of transfer learning is overcoming parameter overfit to the source task; when the parameters have not fully converged, the network as a whole exhibits more plasticity, allowing for faster adaptation. 
We constructed this metric to penalize methods that performed well on the target task only due to underperforming on the source task. 
For all of these metrics, we calculate the mean and standard deviation of a bootstrapped sampling the runs of each method, and calculated the interquartile mean (IQM) and the bootstrapped 95\% confidence interval per~\citet{agarwal_deep_2021}. 

One of the key assumptions that we make in the motivation of this work is that in real-world online task transfer scenarios, the policy is assumed to have converged to maximize the performance on the source task before novelty is injected and the policy must be adapted to the target task. 
However, in practice one of the deficiencies of deep reinforcement learning is the highly stochastic nature of convergence, especially in sparse reward tasks like those of NovGrid. 
For an analysis best aligned with our motivations, we measure our results with respect to the full set of experiments that converged on the source task unless otherwise specified.


\section{Results and Discussion}

\begin{figure}
  \centering
  \begin{subfigure}[b]{0.45\textwidth}
    \includegraphics[width=\textwidth]{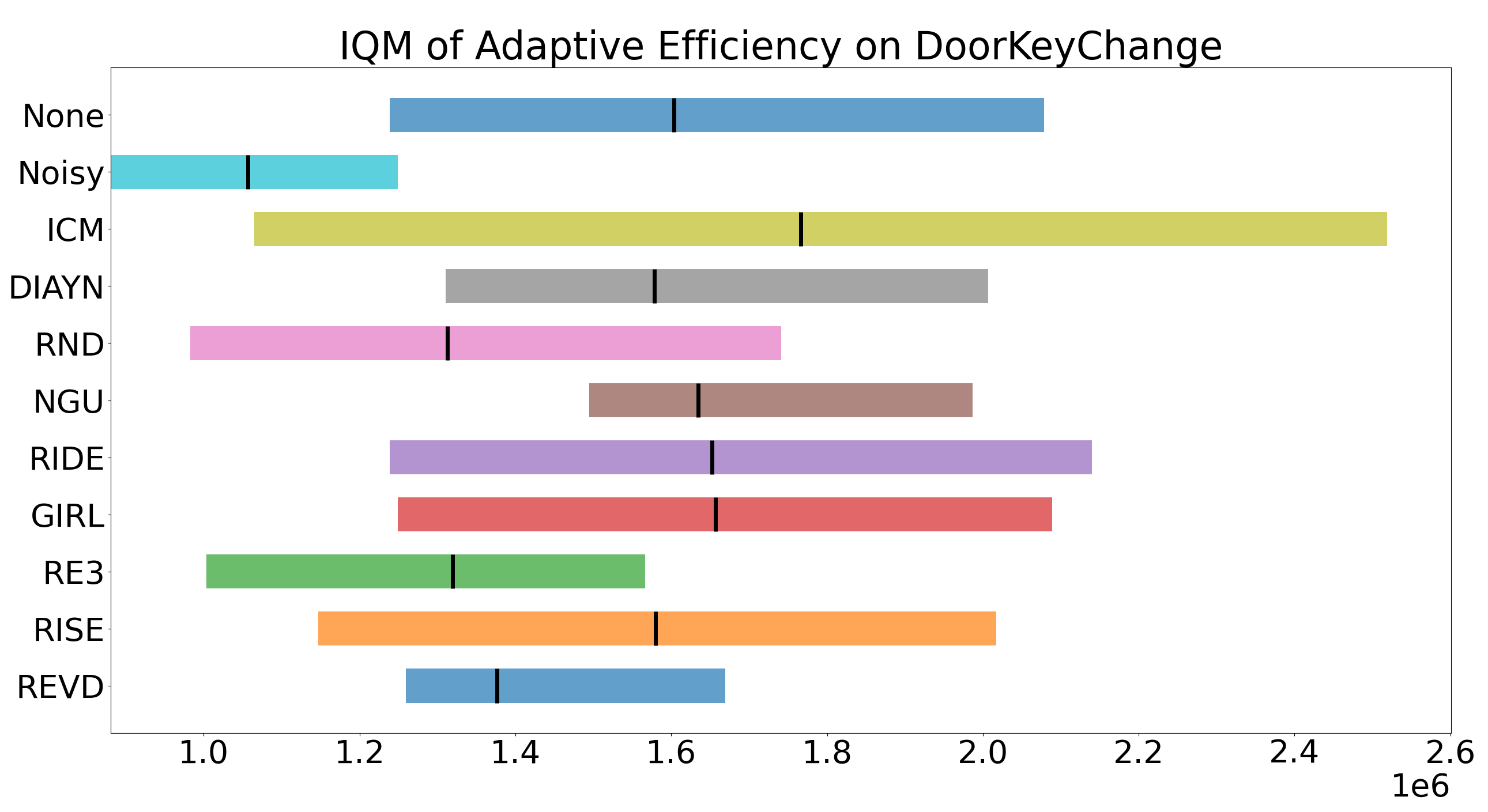}
  \end{subfigure}
  \begin{subfigure}[b]{0.45\textwidth}
    \includegraphics[width=\textwidth]{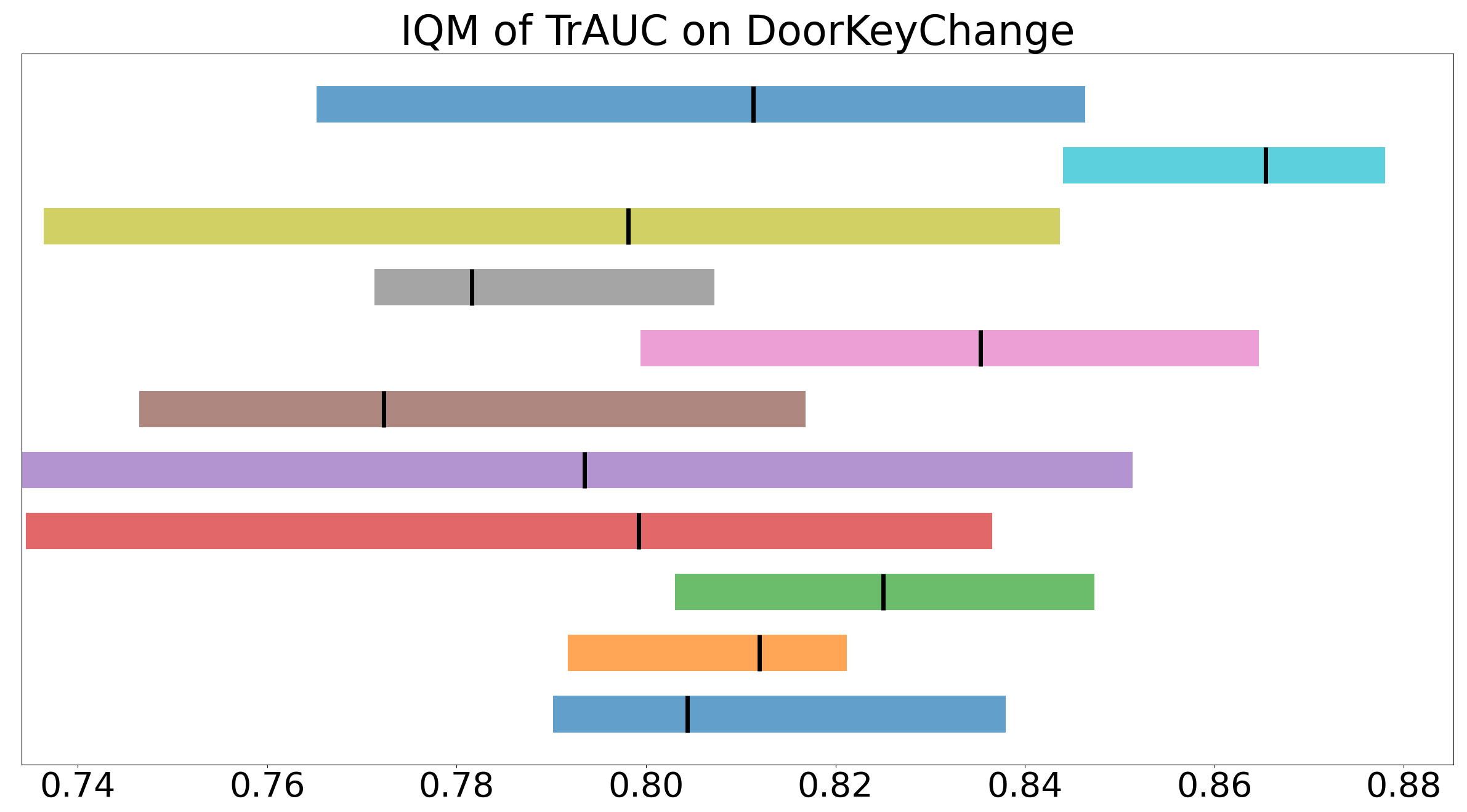}
  \end{subfigure}
  
  \caption{The Adaptive Efficiency and Tr-AUC inter-quartile mean plots for DoorKeyChange. These plots show NoisyNets performing well by both metrics. It should be noted that the Adaptive Efficiency graphs are only showing runs that converged on both tasks and the Tr-AUC graphs are filtering for runs that converged on the first task. 
  }
\end{figure}\label{fig:iqm}




We discuss and analyze our results in the context of our specific experimental research questions:

\paragraph{Are there characteristics of existing exploration algorithms that empirically demonstrate greater suitability for online task transfer overall?}

Across all novelties and environments, we compared the relationship between source task convergence efficiency with adaptive efficiency for different algorithms, and validated our analysis of these comparisons with results on the Tr-AUC metrics, exemplified in Figure~\ref{fig:iqm}. 
A complete list of our results across all algorithms, metrics, environments, novelties can be found in Appendix~\ref{app:results}.
Our analysis shows the following trends hold across most transfer problems: (1) RE3 is the only algorithm that consistently performs well in all metrics tasks and environments, but is only sometimes the best algorithm and outperforms by a very wide margin. 
(2) While NoisyNets, REVD, and to some degree RND have a low convergence efficiency compared to the other exploration methods, on the target task NoisyNets and RE3 have the highest adaptive efficiency. 
(3) ICM, NGU, and other separate objective methods often have high convergence efficiency on the source task but low efficiency on the target task. This is especially true for ICM across most transfer problems.

A pattern we observe is that exploration methods with {\em stochasticity} and {\em explicit diversity} characteristics are slower to converge on the source task, but adapted most efficiently to the target task. 
These methods avoid the trap of inductive bias by using task-agnostic exploration methods. 
However, the impressive performance by NoisyNets on adaptation to the target task, coupled with the poor performance by task-biased methods, suggests that the parameter space learned by NoisyNets is less overfit and therefore better positioned to adapt. 
In the case of REVD and RND,  the relatively slow convergence efficiency on the source task is less noteworthy than the good performance on the target task as both of these methods are designed to use exploration decay in single task learning, which has been removed on all such methods for these transfer tasks. 
This itself compliments our observations about the rapid target task convergence of methods that failed to converge on the source task: for transfer problems in discrete environments there is generally an inverse correlation between source task convergence efficiency and target task adaptive efficiency.

The ICM exploration method is a separate objective exploration method that adds an inductive bias to the typical prediction error-based curiosity metric, focusing on state change predictions \textit{that result from agent action}.
This is a productive approach to single-task learning because it is robust to arbitrary changes in the environment, as described in the ``Noisy TV'' problem~\citep{burda2018exploration}. 
As a result, NGU and several other separate objective algorithms also use this action-focused inductive bias in their embedding spaces. 
In attempting to focus only on action-specific effects, however, this inductive bias reduces likelihood that these agents will efficiently identify and adapt to novelties, which are independent of agent action.

\begin{figure}[t]
    \centering
        \includegraphics[width=0.75\textwidth]{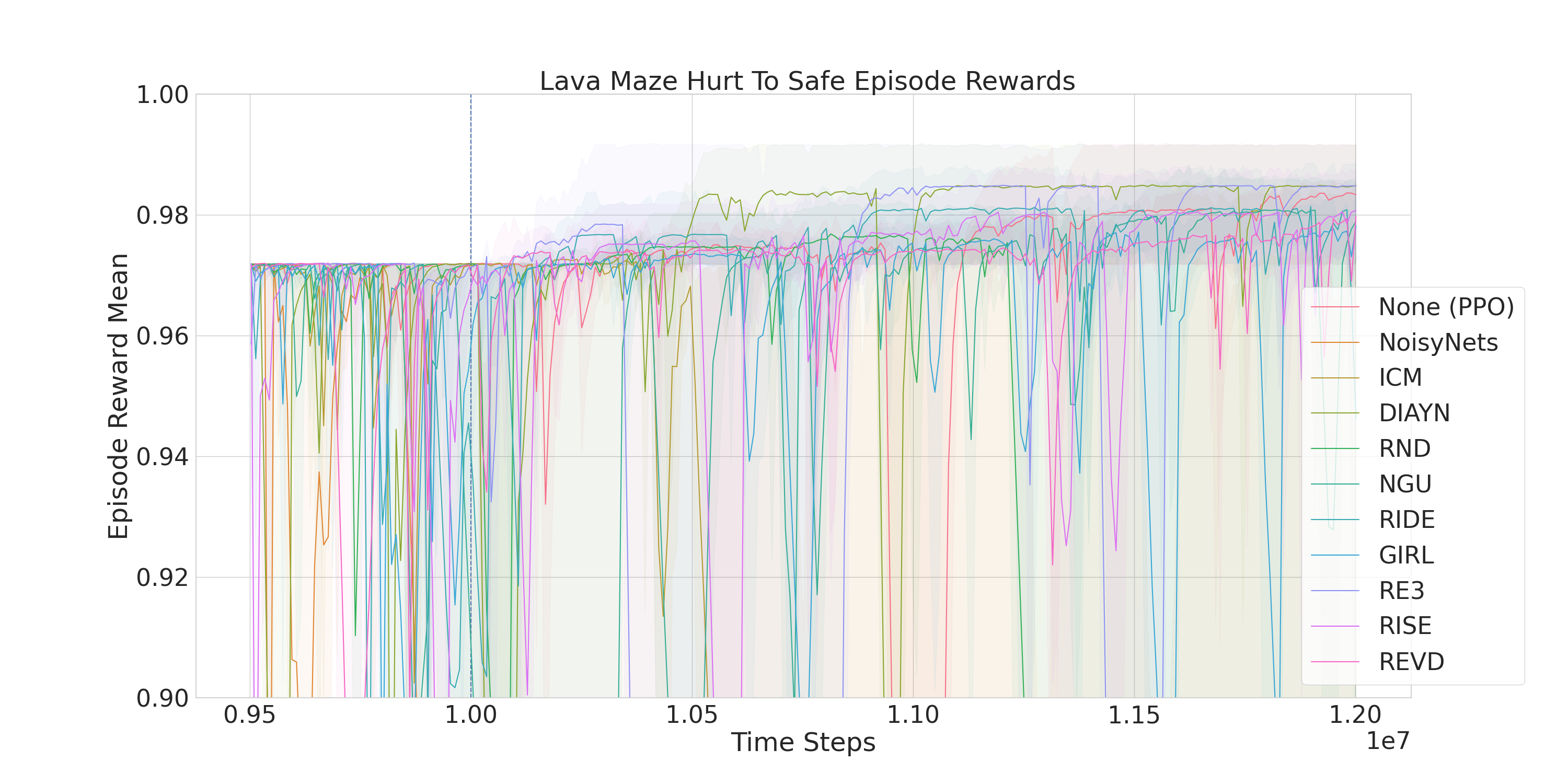}
        \caption{Results from the \texttt{LavaSafe} shortcut novelty. 
        Some of the exploration algorithms are able to find the shortcut, rising above the pre-novelty performance, while others never discover the shortcut. 
        The dotted blue line indicates where novelty was injected.
        }
    \label{fig:walkerandshortcut}
\end{figure}

\begin{figure}[t]
    \centering
    \includegraphics[width=0.75\textwidth]{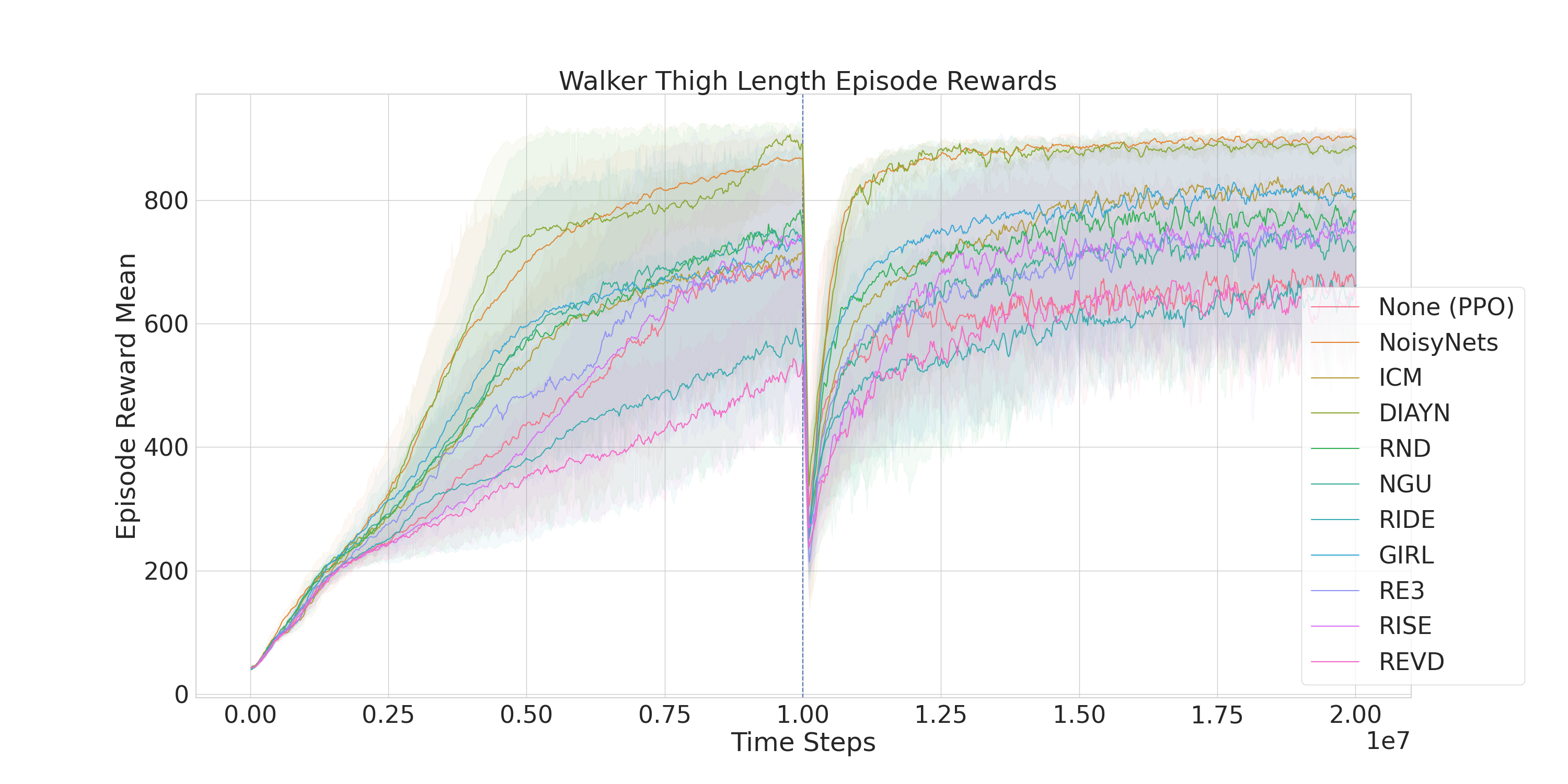}
    \caption{Results from walker thigh length change task.}
    \label{fig:walker}
\end{figure}

\paragraph{How do specific exploration characteristics impact transfer performance in discrete and continuous environments  differently?}

When comparing transfer problems in discrete and continuous environments, we observed two key differences in characteristic importance: (1)~the increased importance of temporal locality, (2)~the decreased importance of explicit diversity. As shown in Figure~\ref{fig:walker}, within the category of temporal locality, we find that the time-independent strategies---NoisyNets and DIAYN---dominate, the temporally global strategies such as RND, ICM, and NGU perform well, and the temporally local strategies struggle the most both pre- and post-novelty. 
Optimal continuous control policies often only need action differences in time. 
As a result, exploration methods that encourage short-term temporal exploration can interfere with continuous control policies for no  dependent models look for diversity. 
Global methods are better at this because they likely notice the global trend that temporal differences and trends are highly complex and perhaps not worth focusing on, and time independent methods are best because there is no risk in getting trapped in trying to maximize exploration with respect to small time horizon changes.

Secondly, continuous control shows a decrease in the importance of diversity in exploration for transfer. 
While some of lack of need of diversity can be attributed to the high transferability of the ThighIncrease novelty, explicit diversity is most useful when the solution does not have many similar actions. 
In discrete control, every action is a unit vector in a basis, so no action is more or less similar to another, and any alternative choice of action adds diversity. 
In discrete control problems, even if repetition is necessary, explicit diversity methods can accommodate because from an information-theoretic standpoint all actions are equally different from one another. 
In continuous control tasks, much of the important knowledge to transfer is the gradual change in control output both before and after 

At the task level beyond the characteristics of the algorithms, one of the biggest differences between novelty adaptation in discrete vs continuous control is the loose correlation between pre- and post-novelty performance. 
While the Tr-AUC metric is motivated by the presumption that poor performance on the source task will lead to deceptively good performance on the target task, in continuous control the opposite is true.
This, as far as we can tell, is evidence that the fundamental knowledge of continuous control is more inherently transferrable. 
While learning to not trip on suddenly long legs does force the agents to forget some of their prior policies. However, much of the challenge in continuous control is learning that at a low-level relationships between action and effect is broadly applicable; moving one joint with an effort of $E$ will be more similar to moving a different joint with the same effort than comparing any two actions in discrete environments. 
The relationship between action and exploration, as we saw in the characteristic analysis, is far more tightly bound than in discrete control.
As a result, inductive biases from separate objectives and controllability assumptions are less problematic, and characteristics that remove time dependence and favor knowledge preservation are more useful.

\paragraph{Do specific exploration characteristics lend themselves well to specific categories of environment change?}

While these observations hold for novelties in general, we do see some differences between the characteristics that favor specific novelties. 
First we compared performance on the DoorKeyChange delta novelty to the CrossingBarrier novelty. 
We see that while RE3, RND, and REVD continue the overarching trend of having comparably high adaptive efficiency, RND and REVD continue to have low convergence efficiency on the source task. 
ICM also continues to have the lowest adaptive efficiency. 
However, the trend with NGU reverses completely: NGU has comparably poor convergence efficiency on the source task, and very good adaptive efficiency post-novelty.

We can make two observations about the difference between delta and barrier novelties that explains the difference in exploration performance. 
First, the CrossingBarrier source task, SimpleCrossing, is a less complex and easier to solve environment than the DoorKey. 
While there is movement among which tasks lead others, there is significantly less variance in the first task convergence, indicating that improved exploration may not be as critical to solving this source task more efficiently.
Second and more importantly, the increased difficulty gap between the source and target task for a barrier novelty means that the transfer learning problem is more similar to single-task learning than a delta novelty.
In transferring from SimpleCrossing to LavaCrossing, the knowledge attained in SimpleCrossing comprises less of the total task understanding required to solve LavaCrossing than changing from one DoorKey to another. 
Therefore, it is not unusual that algorithms that perform well on source tasks in general, like NGU, perform better on barrier novelties than delta and shortcut novelties
%
We can understand from these results that for barriers, diversity methods like RE3 still succeed as they do in general, but that these larger shifts empower methods with well-balanced separate objectives.

The LavaNotSafe barrier novelty is an incredibly challenging novelty for agents to solve at all, but still yields interesting observations. 
The extreme sensitive nature of neural networks to their initialization is well documented by prior work~\citep{hanin2018start}.
The network weights learned in a task as simple as just running directly to a goal and ignoring the environment will in most methods be severely overfit. 
As a result, LavaSafe provides a very poor initialization, even adversarial, initialization for LavaHurts. 
It makes sense then that the stochastic NoisyNets, which avoids overfitting by preserving robust parameters, and explicit diversity methods like RISE, RE3, and REVD perform well on both the Tr-AUC metric and adaptive efficiency in spite of the difficulty of the task. 
In scenarios like this, it is important that exploration algorithms maintain parameters that are easy to unlearn to avoid having an adversarial initialization.


\section{Related Work}\label{relatedwork}

Beyond the specific works mentioned in this research, there is a large body of work characterizing and surveying the impact of exploration on transfer in RL. 
Most of these works focus on proposing a framework for understanding transfer in RL within which exploration is a single element~\citep{taylor2009transfer,Lazaric2012,da2019survey,zhao2020sim,zhu2023transfer} while other focus on exploration with transfer in RL as one of several use cases~\citep{ladosz2022exploration,yang2021exploration}.
There is also a body of work that examines the relationship between active learning and adaptation to novelty and open-worlds~\citep{langley2020open,boult2021towards}, though these works are rarely focused on the particular problems of sequential decision making and RL. 
Our work sets itself apart with deep empirical analysis across a variety of environments and transfer situations. 

There exists research on exploration methods specifically designed for transfer in reinforcement learning (many of which can be found in the aforementioned surveys). 
However, these techniques are usually tailored to a specific algorithm~\citep{zhan2015online,barreto2017successor} or transfer problem in RL, do not directly translate to deep RL~\citep{konidaris2012transfer}, or do not compare themselves to stationary MDP exploration methods. 
This is an important path for research, but separate from the focus of our work investigating standard exploration methods on a variety of transfer problems.
Most similar to our work is the work of \citep{burda2018large}, which empirically investigates the implications of different exploration algorithms that share a curiosity objective as their exploration principle. 
However, our work distinguishes itself by including intrinsic reward within a broader group of exploration principles and algorithmic instantiations for the purposes of transfer in RL.

\section{Conclusions}

In this work we evaluated several deep reinforcement learning exploration algorithms on a variety of online task transfer problems. 
In seeking to understand which attributes of traditional exploration algorithms are important for efficient transfer in RL, our results and analysis reveal four key findings:
(1)~Exploration principles of \textit{explicit diversity}, like RE3, and \textit{stochasticity}, like NoisyNets, are the most consistently positive exploration characteristics across a variety of novelty and environment types. 
(2)~Source task convergence efficiency in discrete control tasks is inversely correlated with target task adaptive efficiency, while the opposite can be true in continuous control. 
(3)~Time-dependent exploration methods, especially short-term \textit{local} temporal locality characteristics, are more poorly suited to transfer in continuous control than in discrete control. 
(4)~The relative importance of exploration characteristics like explicit diversity varies with novelty type.

Moving forward, the research community can use the findings in this work to improve online task transfer in a variety of applications. 
Now, with prior knowledge about changes in the environment that agent RL may face, researchers can select an exploration algorithm well suited to that particular environment and novelty. 
Beyond that, the results of this work will better enable researchers to combine principles and other characteristics of exploration methods; with the knowledge that certain exploration algorithms are well suited to specific transfer situations, future exploration algorithms transfer may be able to dynamically adapt its characteristics for any situation.

\subsubsection*{Broader Impact Statement}
\label{sec:broaderImpact}
As reinforcement learning finds broader, real world applications, we as a research community must understand that improvements for algorithms both increase the likelihood of positive adoption and misuse. 
As this work is motivated by the real-world problem of non-stationary environment adaptation, we believe that our work has the potential for increasing adoption of both positive and negative applications. 
However, we believe the risk from this research is low: our work, while impactful, has a low technical readiness level, and there are many more steps necessary before the results here can be applied to critical systems such as power distribution control.



\bibliography{main}

\begin{thebibliography}{47}
\providecommand{\natexlab}[1]{#1}
\providecommand{\url}[1]{\texttt{#1}}
\expandafter\ifx\csname urlstyle\endcsname\relax
  \providecommand{\doi}[1]{doi: #1}\else
  \providecommand{\doi}{doi: \begingroup \urlstyle{rm}\Url}\fi

\bibitem[Agarwal et~al.(2021)Agarwal, Schwarzer, Castro, Courville, and Bellemare]{agarwal_deep_2021}
Rishabh Agarwal, Max Schwarzer, Pablo~Samuel Castro, Aaron~C Courville, and Marc Bellemare.
\newblock Deep reinforcement learning at the edge of the statistical precipice.
\newblock In \emph{Advances in Neural Information Processing Systems}, volume~34, pp.\  29304--29320. Curran Associates, Inc., 2021.
\newblock URL \url{https://proceedings.neurips.cc/paper/2021/hash/f514cec81cb148559cf475e7426eed5e-Abstract.html}.

\bibitem[Andrychowicz et~al.(2017)Andrychowicz, Wolski, Ray, Schneider, Fong, Welinder, McGrew, Tobin, Pieter~Abbeel, and Zaremba]{andrychowicz2017her}
Marcin Andrychowicz, Filip Wolski, Alex Ray, Jonas Schneider, Rachel Fong, Peter Welinder, Bob McGrew, Josh Tobin, OpenAI Pieter~Abbeel, and Wojciech Zaremba.
\newblock Hindsight experience replay.
\newblock In I.~Guyon, U.~Von Luxburg, S.~Bengio, H.~Wallach, R.~Fergus, S.~Vishwanathan, and R.~Garnett (eds.), \emph{Advances in Neural Information Processing Systems}, volume~30. Curran Associates, Inc., 2017.

\bibitem[Badia et~al.(2020)Badia, Sprechmann, Vitvitskyi, Guo, Piot, Kapturowski, Tieleman, Arjovsky, Pritzel, Bolt, and Blundell]{Badia2020Never}
Adrià~Puigdomènech Badia, Pablo Sprechmann, Alex Vitvitskyi, Daniel Guo, Bilal Piot, Steven Kapturowski, Olivier Tieleman, Martin Arjovsky, Alexander Pritzel, Andrew Bolt, and Charles Blundell.
\newblock Never give up: Learning directed exploration strategies.
\newblock In \emph{International Conference on Learning Representations}, 2020.
\newblock URL \url{https://openreview.net/forum?id=Sye57xStvB}.

\bibitem[Balloch et~al.(2022)Balloch, Lin, Hussain, Srinivas, Peng, Kim, and Riedl]{balloch2022novgrid}
Jonathan Balloch, Zhiyu Lin, Mustafa Hussain, Aarun Srinivas, Xiangyu Peng, Julia Kim, and Mark Riedl.
\newblock Novgrid: A flexible grid world for evaluating agent response to novelty.
\newblock In \emph{In Proceedings of AAAI Symposium, Designing Artificial Intelligence for Open Worlds}, 2022.

\bibitem[Barreto et~al.(2017)Barreto, Dabney, Munos, Hunt, Schaul, van Hasselt, and Silver]{barreto2017successor}
Andr{\'e} Barreto, Will Dabney, R{\'e}mi Munos, Jonathan~J Hunt, Tom Schaul, Hado~P van Hasselt, and David Silver.
\newblock Successor features for transfer in reinforcement learning.
\newblock \emph{Advances in neural information processing systems}, 30, 2017.

\bibitem[Boult et~al.(2021)Boult, Grabowicz, Prijatelj, Stern, Holder, Alspector, Jafarzadeh, Ahmad, Dhamija, Li, et~al.]{boult2021towards}
Terrance Boult, Przemyslaw Grabowicz, Derek Prijatelj, Roni Stern, Lawrence Holder, Joshua Alspector, Mohsen~M Jafarzadeh, Toqueer Ahmad, Akshay Dhamija, Chunchun Li, et~al.
\newblock Towards a unifying framework for formal theories of novelty.
\newblock In \emph{Proceedings of the AAAI Conference on Artificial Intelligence}, volume~35, pp.\  15047--15052, 2021.

\bibitem[Burda et~al.(2018{\natexlab{a}})Burda, Edwards, Pathak, Storkey, Darrell, and Efros]{burda2018large}
Yuri Burda, Harri Edwards, Deepak Pathak, Amos Storkey, Trevor Darrell, and Alexei~A Efros.
\newblock Large-scale study of curiosity-driven learning.
\newblock In \emph{International Conference on Learning Representations}, 2018{\natexlab{a}}.

\bibitem[Burda et~al.(2018{\natexlab{b}})Burda, Edwards, Storkey, and Klimov]{burda2018exploration}
Yuri Burda, Harrison Edwards, Amos Storkey, and Oleg Klimov.
\newblock Exploration by random network distillation.
\newblock In \emph{International Conference on Learning Representations}, 2018{\natexlab{b}}.

\bibitem[Casper et~al.(2023)Casper, Davies, Shi, Gilbert, Scheurer, Rando, Freedman, Korbak, Lindner, Freire, et~al.]{casper2023rlhf}
Stephen Casper, Xander Davies, Claudia Shi, Thomas~Krendl Gilbert, J{\'e}r{\'e}my Scheurer, Javier Rando, Rachel Freedman, Tomasz Korbak, David Lindner, Pedro Freire, et~al.
\newblock Open problems and fundamental limitations of reinforcement learning from human feedback.
\newblock \emph{Transactions on Machine Learning Research}, 2023.

\bibitem[Chentanez et~al.(2004)Chentanez, Barto, and Singh]{chentanez2004intrinsically}
Nuttapong Chentanez, Andrew Barto, and Satinder Singh.
\newblock Intrinsically motivated reinforcement learning.
\newblock \emph{Advances in neural information processing systems}, 17, 2004.

\bibitem[Chevalier-Boisvert et~al.(2023)Chevalier-Boisvert, Dai, Towers, de~Lazcano, Willems, Lahlou, Pal, Castro, and Terry]{MinigridMiniworld23}
Maxime Chevalier-Boisvert, Bolun Dai, Mark Towers, Rodrigo de~Lazcano, Lucas Willems, Salem Lahlou, Suman Pal, Pablo~Samuel Castro, and Jordan Terry.
\newblock Minigrid \& miniworld: Modular \& customizable reinforcement learning environments for goal-oriented tasks.
\newblock \emph{CoRR}, abs/2306.13831, 2023.

\bibitem[Da~Silva \& Costa(2019)Da~Silva and Costa]{da2019survey}
Felipe~Leno Da~Silva and Anna Helena~Reali Costa.
\newblock A survey on transfer learning for multiagent reinforcement learning systems.
\newblock \emph{Journal of Artificial Intelligence Research}, 64:\penalty0 645--703, 2019.

\bibitem[Dulac-Arnold et~al.(2021)Dulac-Arnold, Levine, Mankowitz, Li, Paduraru, Gowal, and Hester]{Dulac-Arnold2021}
Gabriel Dulac-Arnold, Nir Levine, Daniel~J. Mankowitz, Jerry Li, Cosmin Paduraru, Sven Gowal, and Todd Hester.
\newblock Challenges of real-world reinforcement learning: definitions, benchmarks and analysis.
\newblock \emph{Machine Learning}, 110\penalty0 (9):\penalty0 2419--2468, Sep 2021.
\newblock ISSN 1573-0565.
\newblock \doi{10.1007/s10994-021-05961-4}.
\newblock URL \url{https://doi.org/10.1007/s10994-021-05961-4}.

\bibitem[Eysenbach et~al.(2019)Eysenbach, Gupta, Ibarz, and Levine]{eysenbach2018diversity}
Benjamin Eysenbach, Abhishek Gupta, Julian Ibarz, and Sergey Levine.
\newblock Diversity is all you need: Learning skills without a reward function.
\newblock In \emph{International Conference on Learning Representations}, 2019.
\newblock URL \url{https://openreview.net/forum?id=SJx63jRqFm}.

\bibitem[Haarnoja et~al.(2018)Haarnoja, Zhou, Abbeel, and Levine]{haarnoja2018soft}
Tuomas Haarnoja, Aurick Zhou, Pieter Abbeel, and Sergey Levine.
\newblock Soft actor-critic: Off-policy maximum entropy deep reinforcement learning with a stochastic actor.
\newblock In \emph{International conference on machine learning}, pp.\  1861--1870. PMLR, 2018.

\bibitem[Hanin \& Rolnick(2018)Hanin and Rolnick]{hanin2018start}
Boris Hanin and David Rolnick.
\newblock How to start training: The effect of initialization and architecture.
\newblock \emph{Advances in neural information processing systems}, 31, 2018.

\bibitem[Hazan et~al.(2019)Hazan, Kakade, Singh, and Van~Soest]{hazan2019provably}
Elad Hazan, Sham Kakade, Karan Singh, and Abby Van~Soest.
\newblock Provably efficient maximum entropy exploration.
\newblock In \emph{International Conference on Machine Learning}, pp.\  2681--2691. PMLR, 2019.

\bibitem[Ibarz et~al.(2021)Ibarz, Tan, Finn, Kalakrishnan, Pastor, and Levine]{rlroboticssurvey}
Julian Ibarz, Jie Tan, Chelsea Finn, Mrinal Kalakrishnan, Peter Pastor, and Sergey Levine.
\newblock How to train your robot with deep reinforcement learning: lessons we have learned.
\newblock \emph{The International Journal of Robotics Research}, 40\penalty0 (4-5):\penalty0 698--721, 2021.
\newblock \doi{10.1177/0278364920987859}.
\newblock URL \url{https://doi.org/10.1177/0278364920987859}.

\bibitem[Kiran et~al.(2021)Kiran, Sobh, Talpaert, Mannion, Al~Sallab, Yogamani, and P{\'e}rez]{kiran2021surveydriving}
B~Ravi Kiran, Ibrahim Sobh, Victor Talpaert, Patrick Mannion, Ahmad~A Al~Sallab, Senthil Yogamani, and Patrick P{\'e}rez.
\newblock Deep reinforcement learning for autonomous driving: A survey.
\newblock \emph{IEEE Transactions on Intelligent Transportation Systems}, 23\penalty0 (6):\penalty0 4909--4926, 2021.

\bibitem[Konidaris et~al.(2012)Konidaris, Scheidwasser, and Barto]{konidaris2012transfer}
George Konidaris, Ilya Scheidwasser, and Andrew~G Barto.
\newblock Transfer in reinforcement learning via shared features.
\newblock 2012.

\bibitem[Ladosz et~al.(2022)Ladosz, Weng, Kim, and Oh]{ladosz2022exploration}
Pawel Ladosz, Lilian Weng, Minwoo Kim, and Hyondong Oh.
\newblock Exploration in deep reinforcement learning: A survey.
\newblock \emph{Information Fusion}, 85:\penalty0 1--22, 2022.

\bibitem[Langley(2020)]{langley2020open}
Pat Langley.
\newblock Open-world learning for radically autonomous agents.
\newblock In \emph{Proceedings of the AAAI Conference on Artificial Intelligence}, volume~34, pp.\  13539--13543, 2020.

\bibitem[Langley(2022)]{langley2022agents}
Pat Langley.
\newblock Agents of exploration and discovery.
\newblock \emph{Ai Magazine}, 42\penalty0 (4):\penalty0 72--82, 2022.

\bibitem[Lazaric(2012)]{Lazaric2012}
Alessandro Lazaric.
\newblock \emph{Transfer in Reinforcement Learning: A Framework and a Survey}, pp.\  143--173.
\newblock Springer Berlin Heidelberg, 2012.

\bibitem[Lillicrap et~al.(2019)Lillicrap, Hunt, Pritzel, Heess, Erez, Tassa, Silver, and Wierstra]{lillicrap2019continuous}
Timothy~P. Lillicrap, Jonathan~J. Hunt, Alexander Pritzel, Nicolas Heess, Tom Erez, Yuval Tassa, David Silver, and Daan Wierstra.
\newblock Continuous control with deep reinforcement learning, 2019.

\bibitem[McCloskey \& Cohen(1989)McCloskey and Cohen]{mccloskey1989catastrophic}
Michael McCloskey and Neal~J Cohen.
\newblock Catastrophic interference in connectionist networks: The sequential learning problem.
\newblock In \emph{Psychology of learning and motivation}, volume~24, pp.\  109--165. Elsevier, 1989.

\bibitem[Mnih et~al.(2015)Mnih, Kavukcuoglu, Silver, Rusu, Veness, Bellemare, Graves, Riedmiller, Fidjeland, Ostrovski, et~al.]{mnih2015human}
Volodymyr Mnih, Koray Kavukcuoglu, David Silver, Andrei~A Rusu, Joel Veness, Marc~G Bellemare, Alex Graves, Martin Riedmiller, Andreas~K Fidjeland, Georg Ostrovski, et~al.
\newblock Human-level control through deep reinforcement learning.
\newblock \emph{nature}, 518\penalty0 (7540):\penalty0 529--533, 2015.

\bibitem[Pan \& Yang(2009)Pan and Yang]{pan2009survey}
Sinno~Jialin Pan and Qiang Yang.
\newblock A survey on transfer learning.
\newblock \emph{IEEE Transactions on knowledge and data engineering}, 22\penalty0 (10):\penalty0 1345--1359, 2009.

\bibitem[Pathak et~al.(2017)Pathak, Agrawal, Efros, and Darrell]{pathak2017icm}
Deepak Pathak, Pulkit Agrawal, Alexei~A. Efros, and Trevor Darrell.
\newblock Curiosity-driven exploration by self-supervised prediction.
\newblock In \emph{Proceedings of the 34th International Conference on Machine Learning - Volume 70}, ICML'17, pp.\  2778–2787. JMLR.org, 2017.

\bibitem[Plappert et~al.(2018)Plappert, Houthooft, Dhariwal, Sidor, Chen, Chen, Asfour, Abbeel, and Andrychowicz]{plappert2018parameter}
Matthias Plappert, Rein Houthooft, Prafulla Dhariwal, Szymon Sidor, Richard~Y. Chen, Xi~Chen, Tamim Asfour, Pieter Abbeel, and Marcin Andrychowicz.
\newblock Parameter space noise for exploration.
\newblock In \emph{International Conference on Learning Representations}, 2018.
\newblock URL \url{https://openreview.net/forum?id=ByBAl2eAZ}.

\bibitem[Raffin et~al.(2021)Raffin, Hill, Gleave, Kanervisto, Ernestus, and Dormann]{raffin2021stable}
Antonin Raffin, Ashley Hill, Adam Gleave, Anssi Kanervisto, Maximilian Ernestus, and Noah Dormann.
\newblock Stable-baselines3: Reliable reinforcement learning implementations.
\newblock \emph{Journal of Machine Learning Research}, 22\penalty0 (268):\penalty0 1--8, 2021.

\bibitem[Raileanu \& Rockt{\"a}schel(2019)Raileanu and Rockt{\"a}schel]{raileanu2019ride}
Roberta Raileanu and Tim Rockt{\"a}schel.
\newblock Ride: Rewarding impact-driven exploration for procedurally-generated environments.
\newblock In \emph{International Conference on Learning Representations}, 2019.

\bibitem[Réale et~al.(2007)Réale, Reader, Sol, McDougall, and Dingemanse]{reale2001temperament}
Denis Réale, Simon~M. Reader, Daniel Sol, Peter~T. McDougall, and Niels~J. Dingemanse.
\newblock Integrating animal temperament within ecology and evolution.
\newblock \emph{Biological Reviews}, 82\penalty0 (2):\penalty0 291--318, 2007.
\newblock \doi{https://doi.org/10.1111/j.1469-185X.2007.00010.x}.
\newblock URL \url{https://onlinelibrary.wiley.com/doi/abs/10.1111/j.1469-185X.2007.00010.x}.

\bibitem[Schmidhuber(1991{\natexlab{a}})]{schmidhuber1991curious}
J{\"u}rgen Schmidhuber.
\newblock Curious model-building control systems.
\newblock In \emph{Proc. international joint conference on neural networks}, pp.\  1458--1463, 1991{\natexlab{a}}.

\bibitem[Schmidhuber(1991{\natexlab{b}})]{Schmidhuber91apossibility}
Jürgen Schmidhuber.
\newblock A possibility for implementing curiosity and boredom in model-building neural controllers, 1991{\natexlab{b}}.

\bibitem[Schulman et~al.(2017)Schulman, Wolski, Dhariwal, Radford, and Klimov]{schulman2017proximal}
John Schulman, Filip Wolski, Prafulla Dhariwal, Alec Radford, and Oleg Klimov.
\newblock Proximal policy optimization algorithms.
\newblock \emph{arXiv preprint arXiv:1707.06347}, 2017.

\bibitem[Silver et~al.(2013)Silver, Yang, and Li]{silver2013lifelong}
Daniel~L Silver, Qiang Yang, and Lianghao Li.
\newblock Lifelong machine learning systems: Beyond learning algorithms.
\newblock In \emph{2013 AAAI spring symposium series}, 2013.

\bibitem[Sutton \& Barto(2018)Sutton and Barto]{sutton2018reinforcement}
Richard~S Sutton and Andrew~G Barto.
\newblock \emph{Reinforcement learning: An introduction}.
\newblock MIT press, 2018.

\bibitem[Taylor \& Stone(2009)Taylor and Stone]{taylor2009transfer}
Matthew~E Taylor and Peter Stone.
\newblock Transfer learning for reinforcement learning domains: A survey.
\newblock \emph{Journal of Machine Learning Research}, 10\penalty0 (7), 2009.

\bibitem[Yang et~al.(2021)Yang, Tang, Bai, Liu, Hao, Meng, Liu, and Wang]{yang2021exploration}
Tianpei Yang, Hongyao Tang, Chenjia Bai, Jinyi Liu, Jianye Hao, Zhaopeng Meng, Peng Liu, and Zhen Wang.
\newblock Exploration in deep reinforcement learning: a comprehensive survey.
\newblock \emph{arXiv preprint arXiv:2109.06668}, 2021.

\bibitem[Yu et~al.(2020)Yu, Lyu, and Tsang]{yu20girl}
Xingrui Yu, Yueming Lyu, and Ivor Tsang.
\newblock Intrinsic reward driven imitation learning via generative model.
\newblock In Hal~Daumé III and Aarti Singh (eds.), \emph{Proceedings of the 37th International Conference on Machine Learning}, volume 119 of \emph{Proceedings of Machine Learning Research}, pp.\  10925--10935. PMLR, 13--18 Jul 2020.
\newblock URL \url{https://proceedings.mlr.press/v119/yu20d.html}.

\bibitem[Yuan et~al.(2022{\natexlab{a}})Yuan, Li, Jin, and Zeng]{yuan2022rewarding}
Mingqi Yuan, Bo~Li, Xin Jin, and Wenjun Zeng.
\newblock Rewarding episodic visitation discrepancy for exploration in reinforcement learning.
\newblock In \emph{Deep Reinforcement Learning Workshop NeurIPS 2022}, 2022{\natexlab{a}}.

\bibitem[Yuan et~al.(2022{\natexlab{b}})Yuan, Pun, and Wang]{yuan2022renyi}
Mingqi Yuan, Man-On Pun, and Dong Wang.
\newblock R{\'e}nyi state entropy maximization for exploration acceleration in reinforcement learning.
\newblock \emph{IEEE Transactions on Artificial Intelligence}, 2022{\natexlab{b}}.

\bibitem[Zhan \& Taylor(2015)Zhan and Taylor]{zhan2015online}
Yusen Zhan and Mattew~E Taylor.
\newblock Online transfer learning in reinforcement learning domains.
\newblock In \emph{2015 AAAI Fall Symposium Series}, 2015.

\bibitem[Zhang et~al.(2019)Zhang, Zhang, and Qiu]{powerrlsurvey}
Zidong Zhang, Dongxia Zhang, and Robert~C Qiu.
\newblock Deep reinforcement learning for power system applications: An overview.
\newblock \emph{CSEE Journal of Power and Energy Systems}, 6\penalty0 (1):\penalty0 213--225, 2019.

\bibitem[Zhao et~al.(2020)Zhao, Queralta, and Westerlund]{zhao2020sim}
Wenshuai Zhao, Jorge~Pe{\~n}a Queralta, and Tomi Westerlund.
\newblock Sim-to-real transfer in deep reinforcement learning for robotics: a survey.
\newblock In \emph{2020 IEEE symposium series on computational intelligence (SSCI)}, pp.\  737--744. IEEE, 2020.

\bibitem[Zhu et~al.(2023)Zhu, Lin, Jain, and Zhou]{zhu2023transfer}
Zhuangdi Zhu, Kaixiang Lin, Anil~K Jain, and Jiayu Zhou.
\newblock Transfer learning in deep reinforcement learning: A survey.
\newblock \emph{IEEE Transactions on Pattern Analysis and Machine Intelligence}, 2023.

\end{thebibliography}
\bibliographystyle{rlc}
\newpage
\appendix

\section{Algorithm Descriptions}\label{app:algos}

\textbf{RND:} Random Network Distillation is an exploration algorithm that uses the error of a randomly generated prediction problem as an intrinsic reward for the agent. The prediction problem is set up with two neural networks: a randomly initialized fixed target network and a predictor network that is attempting to approximate the target network. Both networks take an observation and output a $k$-dimensional latent vector. The predictor network is trained on observations collected from the agent using gradient descent to minimize the MSE between the outputs of the two neural networks. This MSE loss is used as the intrinsic reward, which will be higher when the predictor network and target network have not been trained on an observation enough to learn the latent yet.

\textbf{REVD:} Rewarding Episodic Visitation Discrepancy is an exploration method that uses intrinsic rewards to motivate the agent to maximize the discrepancy between the set of states visited in consecutive episodes. The discrepancy between consecutive episodes is measured by an estimate of the Renyi divergence using samples from the two episodes. The intrinsic reward is calculated by using the term in the divergence estimate that has to do with the current state, incentivising the agent to visit states that will increase the divergence estimate between the current episode and the previous one.

\textbf{RE3:} Random Encoders for Efficient Exploration is an exploration method that sets the intrinsic reward to an estimate of state entropy. To estimate state entropy, the method applies a k-nearest neighbor entropy estimator in a low-dimensional space the observations are mapped to using a randomly initialized fixed convolutional encoder. The encoder does not need to be trained and instead relies on the convolutional structure of the network, making the algorithm computationally efficient.

\textbf{RIDE:} Rewarding Impact-Driven Exploration is an exploration method that uses intrinsic rewards to incentivize the agent to take actions that lead to large changes in a learned state representation. The learned state representation comes from an encoder that allows for learning of both the forward and inverse models (taken from ICM). The learning problems the state representation is used for only incentivizes the encoder to retain features of the environment that are influenceable by the agent’s actions. Thus, the intrinsic reward is defined as the difference in said state representation, allowing the agent to experience a diverse set of states.

\textbf{ICM:} Intrinsic Curiosity Module is an exploration method that uses the prediction error of a forward model that acts on state embeddings as the intrinsic reward. The state embeddings are learned by using these embeddings to learn an inverse model to predict the action that takes a state embedding to the state embedding in the next time step. These state embeddings are learned to only contain information relevant to the inverse model, effectively solving the noisy-tv problem. The prediction error of the forward model as an intrinsic reward motivates the agent to explore states that it has a poor estimate of the forward dynamics, which should correlate with states the agent has observed less.

\textbf{NGU:} Never Give Up is an exploration algorithm that constructs an intrinsic reward to strongly discourage revisiting the same state more than once within an episode and discourage visiting states that have been visited many times before. These goals are achieved by an episodic novelty module and a life-long novelty module respectively. These use the embedding networks trained in the same manner as ICM to generate a meaningful lower dimensional state representation. The episodic novelty module uses episodic memory and a k-nearest neighbors pseudo-count method to calculate the intrinsic reward. The life-long novelty module uses the same method as RND. Then these two values are combined using multiplicative modulation for the final intrinsic reward.

\textbf{NoisyNets:} Noisy Networks is an exploration algorithm that applies parametric noise to the weights to introduce stochasticity in the agent's policy. This method adds very little overhead since all it requires is a few extra noise parameters in a few layers of the network. This added stochasticity in the weights propagates to the agent's policy to lead to the agent exploring more unknown states instead of only acting greedily.

\textbf{GIRL:} Generative Intrinsic Reward Learning is an exploration algorithm that motivates the agent to visit areas in which a separate model attempting to model the conditional state distribution performs poorly. The method does this by adding an intrinsic reward of the reconstruction error of each state to the extrinsic reward from the task. The model used to model the state distribution is a conditional VAE conditioned on the previous state and a latent variable.

\textbf{RISE:} Renyi State Entropy Maximization is an exploration algorithm that uses intrinsic rewards to maximize the estimate of intra episode Renyi state entropy. This estimate is calculated on latent embeddings of the states within an episode, where the latents are taken from a VAE trained to reconstruct the states. Further, the algorithm automatically searches the different possibilities for the value of k used in the KNN for the Renyi state entropy estimation that guarantees estimation accuracy. Lastly, RISE uses the distance between each state and its k-nearest neighbors as an estimate for entropy and sets the intrinsic reward to this value. The goal of this reward is to motivate the agent to visit a diverse set of states that increases the entropy of the agent’s state visitations. This method is computationally efficient and does not require any additional memory or networks to backpropagate through.

\textbf{DIAYN:} Diversity Is All You Need is an exploration pre-training method that learns a skill-conditioned policy with the goal to produce diverse skills. This is done by setting the reward to something correlated with the performance of a discriminator model that attempts to predict the skill by using the current state as input. Each episode a new skill is sampled for the policy to use, and the discriminator must attempt to predict the skill. Theoretically, this should lead to the policy attempting to make the job of the discriminator as easy as possible by creating diverse skills. Note that in the original paper this reward and skill-conditioned policy was used before any task reward was introduced. Then, these diverse skills were used to learn a task. However, in our work, we adapt DIAYN to be an online algorithm where this reward is trained simultaneously with the task reward. This motivates the agent to both solve the task while keeping the discriminator's job easy by ensuring different skills cover different areas of the state space. This online adaptation of DIAYN works as a traditional exploration algorithm by motivating the agent to take diverse paths throughout training by sampling different diverse skills to use each episode.

\subsection{A note on ``online'' DIAYN}

The effectiveness of explicit diversity and stochasticity methods is consistent throughout our results; however, this does not mean that adding diversity or stochasticity to any algorithm in any way will guarantee improvement to that algorithm's efficiency in novelty adaptation. 
The fundamental design of an algorithm to succeed in a specific RL problem, such as online task transfer, is as important as the selection of exploration principle and instantiation. For example, online DIAYN has average efficiency in both pre and post-novelty for all tasks we tested it on. However, based on the fact that it blends stochasticity with diverse skills could be interpreted to mean that it ought to have performed better post-novelty. In reality, DIAYN's absence of better performance is more likely due to its implementation; as an algorithm originally designed for reward-free pretraining, naive conversion to an online algorithm, while consistent with the original work and able to learn, is a handicap that cannot be solely compensated for by the potential of its exploration approach. A more transfer-appropriate version of DIAYN—as with all of these algorithms—can be designed from scratch and would likely outperform even the best exploration method investigated here. However, this level of algorithmic design ought to be carefully done with the learning problem in mind and is beyond the scope of this work.

\newpage
\section{Exploration Characteristics: Algorithmic Instantiation}\label{app:chars}

We do not report many interesting findings on algorithmic instantiation, partially because our results show that in general algorithmic instantiation does not have an outsized impact on the final results.
While NoisyNets with an update function instantiation is consistently high performing in different transfer problems, so is RE3 using an intrinsic reward. 
Moreover, considering a within-group evaluation of all of the intrinsic reward algorithms, we can see that there is a very high variance over average performance across all metrics; ICM consistently performing poorly, RE3 and REVD consistently performing well, and many of the others performing inconsistently with respect to one another.
Maybe most critically, however, we do not think it wise to generalize over conclusions about algorithmic instantiation from this work because of all of the characteristic categories, algorithmic instantiation is the most unbalanced. 
The vast majority of the algorithms evaluated in this paper are intrinsic reward, while only one, NoisyNets, has a modified update function, and even DIAYN, while altering the environment sampling process by a policy conditioned on a random skill vector, still uses an intrinsic reward as well. 
This imbalance is accidental, but not unexpected; the vast majority of modern exploration algorithm that generalize to different problems like we used here use intrinsic reward. 
An important direction of future work will be to construct fair means of comparison with offline algorithms and algorithms only suited for continuous control or discrete control so that more methods like $\epsilon$-greedy~\citep{sutton2018reinforcement}, maximum entropy RL~\citep{hazan2019provably, haarnoja2018soft}, and replay methods like hindsight experience replay~\cite{andrychowicz2017her} can also be compared.

\newpage
\section{Additional Analysis}\label{app:analysis}

We also examined the shortcut LavaProof novelty as compared to the other novelties, and we see some interesting behavior very specific to the notion of a shortcut. 
As identified in prior work, shortcuts can be notoriously hard exploration problems for transfer learning because the novelty is injected and the learner's prior optimum is undisturbed. 
As we have noted, if we used exploration decay in our algorithm implementations, as is common in single-task RL, there is a chance most or even all of the algorithms in this study would ignore the new shortcut and continue with the sub-optimal solution. 
Even without exploration decay, NGU, GIRL, and ICM all fail to consistently identify the shortcut over the safe lava in spite of learning how to safely navigate around it. 
Atypically, NoisyNets also performed poorly and was unable to consistently find the novelty. 
Of those that performed well, in addition to RE3, DIAYN and RIDE performed unusually well. 
These observations together serve as strong evidence that the main difference in characteristic importance for shortcuts is an even stronger emphasis on the importance of explicit diversity. 
For a shortcut, the critical steps are to (1) identify that a shortcut exists, and (2) consider it worth exploring. 
Although intuitively the stochastic nature of NoisyNets may thrive at shortcut identification, it is less likely that a time-independent method like NoisyNets would be able to value exploring something just because it was novel. 
In this way, the lack of temporal locality in NoisyNets overcomes its potential for exploring the novelty.
Interestingly, the reverse happens for DIAYN. 
DIAYN's core motivation is to learn separable distinguishable policy skills, which for a single task learning problem becomes progressively harder as the policy converges. 
When a shortcut is identified, there is a novel opportunity for DIAYN to suddenly learn more diverse separable skills. 
As a result, the DIAYN's specific implementation of explicit diversity is able to overcome its time-independent exploration nature.

\newpage
\section{Additional Results}\label{app:results}

\begin{table}[ht]
    \centering
    \footnotesize
    \begin{tabular}{|c|c|c|c|c|c|}
        \hline
          & \multicolumn{5}{|c|}{Convergence Efficiency $\downarrow$} \\ 
         \cline{2-6} 
        Exploration & DoorKeyChange & LavaNotSafe & LavaProof & CrossingBarrier & ThighIncrease \\ 
        Algorithm & ($10^{6}$) & ($10^{5}$) & ($10^{6}$) & ($10^{5}$) & ($10^{6}$) \\ \hline 
        \hline
        None (PPO) & 2.56 $\pm$ 0.584 & 0.707 $\pm$ 0.35 & 1.7 $\pm$ 0.683 & 5.43 $\pm$ 1.69 & 7.99 $\pm$ 1.09 \\
        \hline
        NoisyNets & 2.45 $\pm$ 0.908 & 1.02 $\pm$ 0.911 & 1.31 $\pm$ 1.14 & 4.92 $\pm$ 2.13 & 7.17 $\pm$ 1.72 \\
        ICM & 2.12 $\pm$ 0.595 & 0.604 $\pm$ 0.0966 & 1.8 $\pm$ 1.46 & 4.66 $\pm$ 1.14 & 7.34 $\pm$ 1.02 \\
        DIAYN & 2.19 $\pm$ 0.808 & 0.707 $\pm$ 0.265 & 3.44 $\pm$ 1.57 & 5.47 $\pm$ 2.37 & \textbf{6.87 $\pm$ 2.41} \\
        RND & 2.41 $\pm$ 0.956 & 0.635 $\pm$ 0.0893 & 0.976 $\pm$ 0.803 & 5.11 $\pm$ 0.95 & 7.5 $\pm$ 2.04 \\
        NGU & 2.14 $\pm$ 0.289 & 0.768 $\pm$ 0.291 & 2.34 $\pm$ 3.38 & 5.43 $\pm$ 2.03 & 7.72 $\pm$ 1.52 \\
        RIDE & 2.39 $\pm$ 0.975 & \textbf{0.563 $\pm$ 0.0687} & \textbf{0.73 $\pm$ 0.293} & 5.65 $\pm$ 2.11 & 8.24 $\pm$ 1.24 \\
        GIRL & 2.4 $\pm$ 0.855 & 0.676 $\pm$ 0.173 & 2.43 $\pm$ 1.69 & 4.63 $\pm$ 0.979 & 7.61 $\pm$ 1.99 \\
        RE3 & 2.14 $\pm$ 0.616 & 0.604 $\pm$ 0.107 & 1.86 $\pm$ 0.669 & 5.42 $\pm$ 1.37 & 7.78 $\pm$ 0.642 \\
        RISE & 2.32 $\pm$ 0.764 & 0.614 $\pm$ 0.145 & 3.14 $\pm$ 1.89 & \textbf{4.29 $\pm$ 0.788} & 8.55 $\pm$ 0.441 \\
        REVD & \textbf{2.12 $\pm$ 0.891} & 0.635 $\pm$ 0.188 & 1.72 $\pm$ 1.66 & 4.8 $\pm$ 1.12 & 8.73 $\pm$ 0.934 \\
        \hline
    \end{tabular}
    \caption{This table shows the convergence efficiency on the pre-novelty task. It is computed by calculating the number of steps from the start of training until convergence on the first task. Thus, lower numbers are better here. Only runs that converged on the first task are taken into account for this metric.}
    \label{tab:convergence_efficiency}
\end{table}

\begin{table}[ht]
    \centering
    \footnotesize
    \begin{tabular}{|c|c|c|c|c|c|}
        \hline
         & \multicolumn{5}{|c|}{Adaptive Efficiency $\downarrow$} \\ \cline{2-6} 
        Exploration & DoorKeyChange & LavaNotSafe & LavaProof & CrossingBarrier & ThighIncrease \\ 
        Algorithm & ($10^{6}$) & ($10^{6}$) & ($10^{4}$) & ($10^{5}$) & ($10^{6}$) \\ \hline 

        \hline
        None (PPO) & 1.5 $\pm$ 0.477 & 2.56 $\pm$ 2.09 & \textbf{2.05 $\pm$ 0.0} & 6.48 $\pm$ 3.15 & 3.4 $\pm$ 2.51 \\
        \hline
        NoisyNets & \textbf{0.965 $\pm$ 0.204} & 0.963 $\pm$ 0.534 & 7.58 $\pm$ 9.15 & 5.88 $\pm$ 3.72 & 1.69 $\pm$ 0.538 \\
        ICM & 1.57 $\pm$ 0.589 & 7.58 $\pm$ 1.26 & \textbf{2.05 $\pm$ 0.0} & 7.69 $\pm$ 4.69 & 4.18 $\pm$ 1.39 \\
        DIAYN & 1.52 $\pm$ 0.422 & 3.65 $\pm$ 2.47 & 5.8 $\pm$ 5.31 & 5.43 $\pm$ 3.71 & \textbf{1.66 $\pm$ 0.389} \\
        RND & 1.23 $\pm$ 0.385 & 4.64 $\pm$ 3.63 & \textbf{2.05 $\pm$ 0.0} & 5.25 $\pm$ 2.39 & 2.81 $\pm$ 1.46 \\
        NGU & 1.58 $\pm$ 0.317 & 2.39 $\pm$ 1.38 & 6.4 $\pm$ 11.5 & 4.41 $\pm$ 4.02 & 3.71 $\pm$ 1.74 \\
        RIDE & 1.53 $\pm$ 0.527 & 4.51 $\pm$ 2.34 & 2.56 $\pm$ 1.35 & 5.32 $\pm$ 3.71 & 5.18 $\pm$ 2.73 \\
        GIRL & 1.57 $\pm$ 0.541 & 5.49 $\pm$ 3.1 & \textbf{2.05 $\pm$ 0.0} & 6.31 $\pm$ 4.57 & 3.08 $\pm$ 1.98 \\
        RE3 & 1.21 $\pm$ 0.312 & \textbf{0.896 $\pm$ 0.21} & \textbf{2.05 $\pm$ 0.0} & \textbf{4.14 $\pm$ 2.07} & 4.32 $\pm$ 1.81 \\
        RISE & 1.41 $\pm$ 0.374 & 1.37 $\pm$ 0.478 & \textbf{2.05 $\pm$ 0.0} & 4.67 $\pm$ 3.26 & 3.6 $\pm$ 0.597 \\
        REVD & 1.27 $\pm$ 0.319 & 2.43 $\pm$ 1.34 & 2.87 $\pm$ 1.64 & 5.43 $\pm$ 3.24 & 3.92 $\pm$ 0.202 \\
        \hline
    \end{tabular}
    \caption{This table shows the adaptive efficiency on the post-novelty task. it is computed by calculating the number of steps from the start of the novel task until convergence on the second task. Thus, lower numbers are better. Only runs that converged on both tasks are taken into account for this metric.}
    \label{tab:adaptive_efficiency}
\end{table}

\begin{table}[ht]
    \centering
    \footnotesize
    \begin{tabular}{|c|c|c|c|c|c|}
        \hline
         & \multicolumn{5}{|c|}{Transfer Area Under Curve $\uparrow$} \\ \cline{2-6} 
        Exploration & DoorKeyChange & LavaNotSafe & LavaProof & CrossingBarrier & ThighIncrease \\ 
        Algorithm & ($10^{-1}$) & ($10^{-1}$) & ($10^{-1}$) & ($10^{-1}$) & ($10^{2}$) \\ \hline 
        \hline
        None (PPO) & 7.72 $\pm$ 0.792 & 7.43 $\pm$ 1.29 & 9.66 $\pm$ 0.0835 & 8.89 $\pm$ 0.297 & 6.5 $\pm$ 1.63 \\
        \hline
        NoisyNets & \textbf{8.13 $\pm$ 1.23} & \textbf{8.37 $\pm$ 0.885} & 7.69 $\pm$ 3.36 & 8.94 $\pm$ 0.388 & 8.62 $\pm$ 0.39 \\
        ICM & 7.28 $\pm$ 1.07 & 5.43 $\pm$ 0.667 & 9.22 $\pm$ 1.16 & 8.74 $\pm$ 0.537 & 7.25 $\pm$ 1.56 \\
        DIAYN & 7.54 $\pm$ 0.624 & 6.25 $\pm$ 1.22 & \textbf{9.7 $\pm$ 0.0773} & 9.01 $\pm$ 0.493 & \textbf{8.72 $\pm$ 0.203} \\
        RND & 8.09 $\pm$ 0.542 & 6.25 $\pm$ 1.53 & 9.37 $\pm$ 0.66 & 9.0 $\pm$ 0.399 & 7.47 $\pm$ 1.73 \\
        NGU & 7.56 $\pm$ 0.508 & 6.86 $\pm$ 1.38 & 9.48 $\pm$ 0.38 & 9.09 $\pm$ 0.444 & 7.07 $\pm$ 1.68 \\
        RIDE & 7.67 $\pm$ 0.727 & 7.63 $\pm$ 0.895 & 9.5 $\pm$ 0.605 & 9.02 $\pm$ 0.373 & 5.76 $\pm$ 1.67 \\
        GIRL & 7.59 $\pm$ 0.855 & 6.01 $\pm$ 1.08 & 9.55 $\pm$ 0.295 & 8.86 $\pm$ 0.51 & 7.45 $\pm$ 1.74 \\
        RE3 & 8.12 $\pm$ 0.387 & 6.82 $\pm$ 1.48 & 9.37 $\pm$ 0.524 & \textbf{9.1 $\pm$ 0.266} & 6.77 $\pm$ 1.99 \\
        RISE & 7.35 $\pm$ 1.08 & 7.07 $\pm$ 1.77 & 9.42 $\pm$ 0.402 & 9.09 $\pm$ 0.343 & 7.05 $\pm$ 1.03 \\
        REVD & 7.99 $\pm$ 0.402 & 7.3 $\pm$ 1.68 & 9.69 $\pm$ 0.056 & 8.92 $\pm$ 0.384 & 5.58 $\pm$ 1.33 \\
        \hline
    \end{tabular}
    \caption{This table shows the transfer area under the curve metric, which is computed by adding final reward on the first task with the area under the reward curve in the second task. Higher numbers are better here. This only includes runs that converged on the first task.}
    \label{tab:trauc}
\end{table}

\begin{table}[ht]
    \centering
    \footnotesize
    \begin{tabular}{|c|c|c|c|c|c|}
        \hline
         & \multicolumn{5}{|c|}{Adaptive Freq $\uparrow$} \\ \cline{2-6} 
        Exploration & DoorKeyChange & LavaNotSafe & LavaProof & CrossingBarrier & ThighIncrease \\ 
        Algorithm &  & ($10^{-1}$) &  &  &  \\ \hline 

        \hline
        None (PPO) & \textbf{1.0 $\pm$ 0.0} & 6.0 $\pm$ 4.9 & \textbf{1.0 $\pm$ 0.0} & \textbf{1.0 $\pm$ 0.0} & \textbf{1.0 $\pm$ 0.0} \\
        \hline
        NoisyNets & 0.889 $\pm$ 0.314 & \textbf{8.0 $\pm$ 4.0} & 0.714 $\pm$ 0.452 & \textbf{1.0 $\pm$ 0.0} & \textbf{1.0 $\pm$ 0.0} \\
        ICM & 0.889 $\pm$ 0.314 & 3.0 $\pm$ 4.58 & 0.875 $\pm$ 0.331 & \textbf{1.0 $\pm$ 0.0} & \textbf{1.0 $\pm$ 0.0} \\
        DIAYN & \textbf{1.0 $\pm$ 0.0} & 3.0 $\pm$ 4.58 & \textbf{1.0 $\pm$ 0.0} & \textbf{1.0 $\pm$ 0.0} & \textbf{1.0 $\pm$ 0.0} \\
        RND & \textbf{1.0 $\pm$ 0.0} & 3.0 $\pm$ 4.58 & 0.714 $\pm$ 0.452 & \textbf{1.0 $\pm$ 0.0} & \textbf{1.0 $\pm$ 0.0} \\
        NGU & \textbf{1.0 $\pm$ 0.0} & 4.0 $\pm$ 4.9 & \textbf{1.0 $\pm$ 0.0} & 0.9 $\pm$ 0.3 & \textbf{1.0 $\pm$ 0.0} \\
        RIDE & \textbf{1.0 $\pm$ 0.0} & 6.0 $\pm$ 4.9 & \textbf{1.0 $\pm$ 0.0} & \textbf{1.0 $\pm$ 0.0} & \textbf{1.0 $\pm$ 0.0} \\
        GIRL & \textbf{1.0 $\pm$ 0.0} & 2.0 $\pm$ 4.0 & \textbf{1.0 $\pm$ 0.0} & \textbf{1.0 $\pm$ 0.0} & \textbf{1.0 $\pm$ 0.0} \\
        RE3 & \textbf{1.0 $\pm$ 0.0} & 2.0 $\pm$ 4.0 & \textbf{1.0 $\pm$ 0.0} & \textbf{1.0 $\pm$ 0.0} & \textbf{1.0 $\pm$ 0.0} \\
        RISE & 0.857 $\pm$ 0.35 & 3.0 $\pm$ 4.58 & \textbf{1.0 $\pm$ 0.0} & \textbf{1.0 $\pm$ 0.0} & \textbf{1.0 $\pm$ 0.0} \\
        REVD & \textbf{1.0 $\pm$ 0.0} & 5.0 $\pm$ 5.0 & \textbf{1.0 $\pm$ 0.0} & \textbf{1.0 $\pm$ 0.0} & \textbf{1.0 $\pm$ 0.0} \\
        \hline
    \end{tabular}
    \caption{This is the frequency that the agent converges on the second task using this exploration algorithm conditioned on the fast it converged on the first task. Higher numbers are better.}
    \label{tab:adaptive_freq}
\end{table}

\newpage

\section{Implementation Details}
\subsection{Hyperparameters}\label{app:hyperparams}


We sweeped through the hyperparameter configurations for each exploration algorithm using Bayesian hyperparameter optimization. We ran a minimum of 10 hyperparameter configurations (using more for the algorithms with many parameters), each with six runs (three seeds on MiniGrid-DoorKey-8x8-v0 and three seeds on MiniGrid-SimpleCrossingS9N2-v0), for each algorithm. Each successive configuration was calculated using the weights and biases Bayesian sweep method within reasonable preset range around parameters pulled from prior work. The metric optimized for to minimize the average (over the 6 runs) number of steps needed for the \href{https://stable-baselines3.readthedocs.io/en/master/guide/callbacks.html#stoptrainingonrewardthreshold}{StopTrainingOnRewardThreshold} callback from \href{https://stable-baselines3.readthedocs.io/en/master/}{stable-baselines3} to stop the run with a reward threshold set to $0.35$ (capped at 3M steps). Once the sweeps were finished we chose reasonable hyperparameters that followed the trends of the other runs in the sweep to ensure the chosen parameter configuration was not just an outlier.

Here is a table consisting of the ranges of hyperparameters we sweeped through and our final chosen value for them based on the (limited) number of runs we used. The distribution type column refers to the distribution parameter provided to the \href{https://wandb.ai/}{wandb} sweep agent. For specifics about what each parameter does see the individual papers or the implementations in \href{https://github.com/balloch/rl-exploration-transfer/tree/noisy-net-implementation/rlexplore}{our codebase}. Note that latent\_dim, batch\_size, and learning\_rate parameters refer to networks trained specifically for exploration and have nothing to do with the parameters used for policy training.

\begin{table}[ht]
    \centering
    \footnotesize
    \begin{tabular}{|c|c|c|c|c|}
        \hline
        \textbf{Algorithm} & \textbf{Parameter Name} & \textbf{Distribution Type} & \textbf{Range} & \textbf{Final Value} \\
        \hline
        \multirow{1}{*}{\textbf{PPO}} & learning\_rate & q\_uniform & [0.0003, 0.0008] & 0.00075 \\
        \hline
        \multirow{2}{*}{\textbf{RE3}} & beta & q\_log\_uniform\_values & [0.00001, 0.1] & 0.01 \\
        & latent\_dim & categorical & [16, 32, 64, 128, 256] & 64 \\
        \hline
        \multirow{2}{*}{\textbf{RIDE}} & beta & q\_log\_uniform\_values & [0.00001, 0.1] & 0.001 \\
        & latent\_dim & categorical & [16, 32, 64, 128, 256] & 128 \\
        \hline
        \multirow{2}{*}{\textbf{RISE}} & beta & q\_log\_uniform\_values & [0.00001, 0.1] & 0.002 \\
        & latent\_dim & categorical & [16, 32, 64, 128, 256] & 64 \\
        \hline
        \multirow{5}{*}{\textbf{RND}} & beta & q\_log\_uniform\_values & [0.00001, 0.1] & 0.002 \\
        & learning\_rate & q\_log\_uniform\_values & [0.0001, 0.01] & 0.0003 \\
        & batch\_size & categorical & [16, 32, 64] & 64 \\
        & latent\_dim & categorical & [16, 32, 64, 128, 256] & 128 \\
        \hline
        \multirow{1}{*}{\textbf{Noisy Nets}} & num\_noisy\_layers & categorical & [1, 2, 3] & 2 \\
        \hline
        \multirow{5}{*}{\textbf{NGU}} & beta & q\_log\_uniform\_values & [0.0001, 0.5] & 0.0005 \\
        & learning\_rate & q\_log\_uniform\_values & [0.0001, 0.01] & 0.0006 \\
        & batch\_size & categorical & [16, 32, 64] & 64 \\
        & latent\_dim & categorical & [16, 32, 64, 128, 256] & 128 \\
        \hline
        \multirow{3}{*}{\textbf{ICM}} & beta & q\_log\_uniform\_values & [0.00001, 0.1] & 0.0003 \\
        & learning\_rate & q\_log\_uniform\_values & [0.0001, 0.01] & 0.0003 \\
        & batch\_size & categorical & [16, 32, 64] & 64 \\
        \hline
        \multirow{4}{*}{\textbf{GIRL}} & beta & q\_log\_uniform\_values & [0.00001, 0.1] & 0.0005 \\
        & learning\_rate & q\_log\_uniform\_values & [0.0001, 0.01] & 0.002 \\
        & lambda & q\_log\_uniform\_values & [0.001, 0.1] & 0.05 \\
        & latent\_dim & categorical & [32, 64, 128] & 64 \\
        \hline
        \multirow{3}{*}{\textbf{REVD}} & beta & q\_log\_uniform\_values & [0.00001, 0.1] & 0.00005 \\
        & latent\_dim & categorical & [16, 32, 64, 128, 256] & 64 \\
        \hline
        \multirow{3}{*}{\textbf{RIDE}} & beta & q\_log\_uniform\_values & [0.00001, 0.1] & 0.001 \\
        & latent\_dim & categorical & [16, 32, 64, 128, 256] & 128 \\
        \hline
        \multirow{3}{*}{\textbf{RISE}} & beta & q\_log\_uniform\_values & [0.00001, 0.1] & 0.002 \\
        & latent\_dim & categorical & [16, 32, 64, 128, 256] & 64 \\
        \hline
    \end{tabular}
    \caption{Hyperparameter Sweeps for Exploration Algorithms.}
    \label{tab:hyperparameters}
\end{table}

For the continuous control task (Walker), we ran a targeted sweep on CartPole, mainly tuning parameters that were important to our results such as beta and other exploration algorithm specific parameters. We used prior work, results from our MiniGrid sweep, and other heuristics to estimate the ranges to sweep for different parameters. The main parameters that changed relative to the table above were the beta's for each algorithm as the reward scale is very different in walker as opposed to any MiniGrid tasks.

\subsection{Experimental Setup}

For a valid comparison, all the experiments were run using PPO with the same PPO hyperparameters (listed below). Further, the experiments use the \href{https://stable-baselines3.readthedocs.io/en/master/guide/custom_policy.html#default-network-architecture}{default MLP policy} network shapes from the stable-baselines3 PPO class for the experiments and any hyperparameters not specified below were left as default. 

\begin{table}[ht]
    \centering
    \begin{tabular}{|c|c|}
        \hline
        \textbf{Parameter} & \textbf{Value} \\
        \hline
        learning\_rate & 0.00075 \\
        n\_steps & 2048 \\
        batch\_size & 256 \\
        n\_epochs & 4 \\
        gamma & 0.99 \\
        gae\_lambda & 0.95 \\
        clip\_range & 0.2 \\
        ent\_coef & 0.01 \\
        vf\_coef & 0.5 \\
        max\_grad\_norm & 0.5 \\
        \hline
    \end{tabular}
    \caption{PPO Configuration}
    \label{tab:ppo_rl_alg_kwargs}
\end{table}

Each experiment on MiniGrid used 10 seeds with 5 parallel environments each to ensure reliable results, logging all results to wandb for future aggregation and analysis.

Each experiment on Walker used 5 seeds with 10 parallel environments.

For each of the environments, we ran the experiments with a number of steps that led to a high convergence rate with the implemented algorithms so fair comparisons between algorithms could be used on the task two results.

\begin{table}[ht]
    \centering
    \footnotesize
    \begin{tabular}{|c|c|c|c|}
        \hline
        \textbf{Environment Name} & \textbf{Pre Novelty Steps} & \textbf{Post Novelty Steps} & \textbf{MiniGrid Size} \\
        \hline
        \textbf{door\_key\_change} & 5M & 3M & 8x8 \\
        \textbf{simple\_to\_lava\_crossing} & 2M & 3M & 9x9 \\
        \textbf{lava\_maze\_safe\_to\_hurt} & 500,000 & 5M & 8x8 \\
        \textbf{lava\_maze\_hurt\_to\_safe} & 5M & 2M & 8x8 \\
        \textbf{walker\_thigh\_length} & 10M & 10M & N/A \\
        \hline
    \end{tabular}
    \caption{Environment Details}
    \label{tab:environment_details}
\end{table}

We used a few observation wrappers on the environments in the experiment to set the observation space to be the flattened observed image (to work with simple MLP policies).


\end{document}